\documentclass[fleqn]{cas-dc}

\usepackage{algorithm2e}
\usepackage[numbers]{natbib}
\usepackage{mathtools}
\usepackage{amsmath}
\usepackage{caption}
\usepackage{subcaption}
\usepackage{chngcntr}
\usepackage{float}

\def\tsc#1{\csdef{#1}{\textsc{\lowercase{#1}}\xspace}}
\tsc{WGM}
\tsc{QE}
\tsc{EP}
\tsc{PMS}
\tsc{BEC}
\tsc{DE}

\begin{document}
\let\WriteBookmarks\relax
\def\floatpagepagefraction{1}
\def\textpagefraction{.001}



\title [mode = title]{Extreme value forecasting using relevance-based data augmentation with deep learning models }                       

 \author[3]{Junru Hua}
\author[3]{Rahul Ahluwalia}

\author[3]{Rohitash Chandra \corref{cor1}} 
\ead{rohitash.chandra@unsw.edu.au}

\affiliation[3]{Transitional Artificial Intelligence Research Group, School of Mathematics and Statistics, UNSW Sydney, Australia}

\cortext[cor1]{Corresponding author}

\begin{abstract}
Data augmentation with \textit{generative adversarial networks} (GANs) has been popular for class imbalance problems, mainly for pattern classification and computer vision-related applications. Extreme value forecasting is a challenging field that has various applications from finance to climate change problems. In this study, we present a data augmentation framework for extreme value forecasting. In this framework, our focus is on forecasting extreme values using deep learning models in combination with data augmentation models such as  GANs and \textit{synthetic minority oversampling technique} (SMOTE). We use deep learning models such as convolutional long short-term memory (Conv-LSTM) and bidirectional long short-term memory (BD-LSTM) networks  for multistep ahead prediction featuring extremes. We investigate which data augmentation models are the most suitable, taking into account the prediction accuracy overall and at extreme regions, along with computational efficiency. We also present novel strategies for incorporating data augmentation, considering extreme values based on a relevance function.  Our results indicate that the SMOTE-based strategy consistently demonstrated superior adaptability, leading to improved performance across both short- and long-horizon forecasts. Conv-LSTM and BD-LSTM exhibit complementary strengths: the former excels in periodic, stable datasets, while the latter performs better in chaotic or non-stationary sequences.

\end{abstract}
 

\maketitle

\section{Introduction}

Extreme value theory (analysis) is the study of problems where there are outliers present that are either really large or small, often called extreme values \cite{haan2006extreme,gomes2015extreme, gilleland2013software}. This is particularly useful in developing models for extreme value forecasting \cite{zhu2017improved}. In most cases, the extreme values are relevant to a problem but are rare and underrepresented in the data. Class imbalance problems refer to problems that have a large difference in the number of data samples between the classes \cite{abd2013review,galar2011review,4667275}. This becomes an issue with conventional machine learning and deep learning models that are not naturally equipped for class-imbalanced problems and have a bias towards the amount of data for the respective class.  Classification problems tackle this problem in a wide range of applications from the detection of fraud phone calls \cite{fawcett1996combining}, oil spills \cite{kubat1998machine}, natural disasters \cite{8228593}, and medical diagnostics of rare diseases \cite{kooi2017large,chudzik2018microaneurysm}. Apart from classification tasks, imbalanced datasets are also an issue in time series forecasting problems \cite{chou2005forecasting,ahmadzadeh2019rare}, where the deep learning models have to predict the extreme values and not just classify them into classes. These models face similar challenges since the model is heavily influenced by the conventional time series data, which is typically referred to as common values, and only minimally influenced by the extreme values due to their low sample size. Hence, a small sample size of extreme values makes it difficult for the model to learn and forecast extreme values. Extreme value forecasting problems have many applications and frequently appear in the areas of weather forecasting \cite{CLOKE2009613,yozgatligil2018extreme} and stock market volatility prediction\cite{zhao2010extreme,poon2003forecasting} 
Forecasting problems typically feature time series (temporal)  data which inherently has temporal dependencies between consecutive data samples. Recurrent Neural Networks (RNNs) have been designed to target sequence modelling \cite{elman1990finding,medsker1999recurrent}, which makes them useful for language modelling tasks \cite{de2015survey,mikolov2011extensions}, and temporal sequences.  The Long Short-Term Memory (LSTM) network \cite{hochreiter1997long} is an enhanced RNN suited for modelling temporal sequences with long time lags that were difficult to train by conventional RNNs \cite{bengio1994learning}. 
 Although Convolutional Neural Networks (CNNs) have also been applied in sequence modelling tasks and shown competitive results in time series prediction \cite{bai2018empirical,yu2015multi,borovykh2017conditional,chandra2021evaluation}, our focus remains on recurrent architectures, particularly LSTM-based models, as they are better aligned with capturing long-term temporal dependencies.
However, despite their success in time series prediction, RNNs and CNNs are not naturally equipped for extreme value forecasting and class imbalance problems.


Data augmentation methods \cite{wen2020time,maharana2022review,yang2022image} have been used with much success in recent decades to combat these problems. Data resampling strategies, in particular have shown particular promise in combating class imbalance problems \cite{garcia2010exploring}. In the area of classification tasks, there has been a multitude of strategies used, including oversampling and undersampling. Oversampling \cite{shelke2017review} typically involves artificially generating samples for the minority class, while undersampling \cite{anand2010approach} reduces the number of samples from the common class. Both methods attempt to equalise the ratio between the classes.  Earlier works involved strategies that utilised oversampling with replacement \cite{Japkowicz2000TheCI,ling1998data} where extra samples were generated by reusing extreme values; however,  such a strategy proved ineffective in minority class recognition.  Synthetic Minority Oversampling Technique (SMOTE) \cite{chawla2002smote} is a  resampling strategy that has yielded significant improvements in the analysis of class imbalance problems. It utilises a combination of oversampling and undersampling methods to level out the imbalance between the classes with the additional condition that the extra samples generated are synthetically created from the datasets rather than just oversampling with replacement. SMOTE has been demonstrated to be an effective way to combat class imbalance in classification problems \cite{fernandez2018smote,pradipta2021smote,9370099}. SMOTE has been extended to the domain of regression problems and time series forecasting known as SMOTE for regression (SMOTE-R) \cite{torgo2013smote} which generalises  SMOTE for regression problems. Further improvements have been demonstrated in SMOTE-R to improve the quality of synthetic samples  \cite{moniz2017resampling}.
 
Generative Adversarial Networks(GANs) \cite{goodfellow2014generative}  have mostly been used for computer vision and image processing \cite{wang2021generative,cao2018recent} and also gained attention in the media for generative arts \cite{shahriar2022gan,gragnaniello2021gan}. However, they can also be used for generating time series  and tabular data \cite{bourou2021review, xu2019modeling} and have been successfully applied for  class imbalance problems \cite{9733348,durgadevi2021generative,8363576,9093842}.
 Sharma et al. \cite{9733348} combined GANs with SMOTE for pattern classification problems based on tabular datasets.  The original GAN framework has been further extended, leading to ExGAN\cite{bhatia2021exgan}  which utilises GAN to generate realistic and extreme samples for a dataset. Furthermore, Wasserstein GAN \cite{arjovsky2017wasserstein} and Bayesian GANs \cite{saatci2017bayesian} also have been developed that have strengths such as combating the mode collapse problem. The mode collapse problem \cite{bau2019seeing} occurs when a GAN over-optimises for specific discriminators, resulting in the output samples being the same or having low variety. Bayesian GANs \cite{chen2018bayesian} and Wasserstein GANs with gradient penalty \cite{gulrajani2017improved} are useful in alleviating this problem. GANs are flexible and easily extensible to a wide range of problems.  As a result, GANs have shown great potential in the use case of extreme forecasting problems.
 
 Although deep learning models differ from statistical approaches, the concepts from extreme value theory are still applicable in class imbalance classification and time series forecasting problems. In the development of models for extreme value problems, the data must be divided into an extreme set and a common set. Although the classes can be used to distinguish between the extreme samples from common samples in class imbalance classification problems \cite{khan2023review,japkowicz2002class}, it is less straightforward for forecasting problems. Extreme value forecasting typically features continuous time series data. One way to determine which samples should be classified as extreme is through a relevance function \cite{ribeiro2011utility} that maps each data sample to a relevance score. A higher relevance score indicates the sample is more extreme. So, by defining a relevance threshold, all the samples with greater relevance than the threshold are labelled as extreme values. The relevance function and relevance threshold for a given application are usually given by an expert in the field. However, there are several ways to create generalised relevance functions that can be applied to any dataset for the sake of testing deep learning models and data augmentation techniques. Ribeiro et al. \cite{ribeiro2011utility}  developed a method that generated a \textit{piecewise cubic hermite interpolating polynomial} (PCHIP) based upon the box statistics for a given dataset. This relevance function maps each data point to a relevance score in the range of 0 to 1, with extreme samples having a higher relevance. Based on the generated polynomial, a relevance threshold can be chosen to partition the data into extreme values and common values. The method is generalisable to any dataset, regardless of the domain and extending this idea into a relevance-based framework will allow for a generalised framework that can be applied to all extreme forecasting problems.
 
In this paper, we present a relevance-based framework that extends the relevance function proposed by Ribeiro et al. \cite{ribeiro2011utility} and employs data augmentation and deep learning methods for forecasting extremes. We evaluate data augmentation methods (SMOTE-R and GANs) for their effectiveness in generating synthetic data samples for extreme values. 
 Our evaluation considers both overall prediction accuracy and, more importantly, the accuracy of extreme value forecasts. To this end, we adopt the Signal Extreme Ratio (SER), a tail-sensitive extension of RMSE originally proposed by Silva et al. \cite{silva2019ser}. The SER metric is specifically designed to capture model performance in the tail regions, making it well-suited for assessing rare and extreme events. Recent studies have further demonstrated its utility in hydrological forecasting, including ensemble quantile-based deep learning frameworks for flood prediction \cite{chandra2024ensemble} and quantile regression approaches for rainfall–runoff uncertainty estimation \cite{weerts2011quantile}.

 The rest of the paper is organised as follows. In Section 2, we provide a background on deep learning and data augmentation. In Section 3, we present the methodology that features the framework utilising and comparing multiple deep learning models. This is followed by E Results in Section 4 and  discussion and Section 5. Finally we conclude the paper in Section 6. 

\section{Background}

\subsection{Deep learning for time series forecasting } 

Time series forecasting has long relied on traditional statistical models such as the Autoregressive Integrated Moving Average (ARIMA) \cite{box2015time,ho1998use,chen2008forecasting}. While effective for linear and stationary data, ARIMA struggles with nonlinearities and high-noise environments commonly observed in real-world applications \cite{petricua2016limitation}. These limitations have motivated a shift towards machine learning and deep learning approaches that can better capture complex temporal dependencies and nonlinear patterns \cite{sezer2020financial,ozbayoglu2020deep}.

In recent decades, various deep learning models have shown superior performance in time series forecasting tasks.  RNNs and LSTM networks have been successfully applied to domains such as energy demand \cite{8404313}, solar irradiance \cite{8266215}, and petroleum production forecasting \cite{sagheer2019time}, consistently outperforming traditional models. Stacked and bidirectional LSTM (BD-LSTM) models have also been used for forecasting COVID-19 trends \cite{shastri2020time}, where Convolutional LSTM (Conv-LSTM) yielded the best performance.

Chandra et al. \cite{chandra2022deep} conducted a comprehensive study applying various LSTM-based architectures for COVID-19 forecasting in India, leveraging multivariate and multi-step recursive strategies. Furthermore, Goel et al. \cite{goel2017r2n2} proposed a Rsidual RNN (R2N2) that combines vector autoregression with RNNs, offering improved multivariate prediction accuracy.
CNNs  have also demonstrated potential for time series applications. For example, CNNs have been used in energy load forecasting \cite{8489399}, financial time series prediction \cite{mehtab2022analysis}, and have shown competitive performance compared to multilayer perceptron networks. Bai et al. \cite{bai2018empirical}  reported that CNNs can outperform RNNs in a diverse range of sequence modelling problems. Extensions such as dilated CNNs \cite{borovykh2018dilated} and semi-dilated CNNs \cite{hussein2021semi} have been applied in conditional forecasting and epileptic seizure prediction, while hybrid CNN-LSTM models have been explored for inventory management \cite{8789957} and gold price forecasting \cite{livieris2020cnn}. Although CNN-based approaches have shown promise, RNN and LSTM variants remain more suitable for capturing long-term temporal dependencies that are critical in extreme value forecasting problems.

 Multi-step time series forecasting, in particular, presents unique challenges due to error accumulation across prediction horizons. Studies such as Chandra \cite{chandra2021evaluation} emphasize that LSTM-based models, including BD-LSTM and Conv-LSTM architectures, provide robust performance for multi-step prediction tasks. This highlights the importance of careful architecture selection and hyperparameter tuning when applying deep learning to real-world forecasting problems.

\subsection{Data Augmentation}   
Data augmentation is widely used for classification problems, especially problems involving image classification.
Typical transformations applied to images include scaling, cropping, flipping, rotating, translating, colour augmentation (change in brightness, contrast, saturation or hue), and other affine transformations \cite{shorten2019survey}. 
Resampling strategies are a popular data augmentation method for class imbalance problems, initially designed for classification problems, but have also been extended to regression problems. The simplest resampling strategy is oversampling with replacement, also known as oversampling by replication, which is highly susceptible to overfitting because it involves concatenating duplicate minority class samples onto the data set \cite{5128907}. SMOTE is a widely used resampling strategy for solving class imbalance problems due to its effectiveness and relative simplicity \cite{10.5555/3241691.3241712}. SMOTE utilises interpolation between samples in the minority class to synthesise new samples. Besides SMOTE, there have been a variety of interpolating methods that have been adapted for different problems, such as regression and forecasting. Adaptive synthetic sampling (ADASYN) is a variation that places more emphasis on minority cases existing in neighbourhoods dominated by majority cases and generates more synthetic data using these particular minority cases since they are harder to learn \cite{adasyn_alejo}. ADASYN has been used for Alzheimer's disease identification, which outperformed other state-of-the-art models \cite{ahmed2022dad}.


\subsection{Data Augmentation for Time Series Data}

Data augmentation for time series differs from data augmentation for classification problems in that both the targets and features have to be synthesised. However, many of the augmentation techniques used for classification problems can be extended for regression and time series problems. SMOTE for regression, also known as SMOTE-R, adapts SMOTE to regression problems by employing a user-defined relevance score function and threshold to identify the ranges of values that are under-represented \cite{torgo2013smote}. Both the target and feature values for synthesised SMOTE-R samples are generated using a weighted average between the seed and neighbour cases.
 
It is common for time series data to exhibit systematic changes in distribution due to hidden contexts that may emerge from external or unknown factors~\cite{schlimmer1986CDNoisy, widmer2022CDContextTracking,agrahari2022CDLit}. The changes in the relationship between the input and the output of a model are referred to as \textit{concept drift} \cite{tsymbal2004problem}. Concept drift describes a shift in the distribution of the target variable conditional on the predictors, whilst the marginal distribution of the predictors remains unchanged \cite{widmer1994CDLearning}. This occurs when hidden contexts responsible for these shifts are not captured within the model. An example is the  Earth's surface temperature time series which is impacted by the season (a recurring concept drift) \cite{stine2009changes}. Furthermore, a time series observing a customer's spending habits may be influenced by the strength of the economy (a gradual concept drift). 
 
Data augmentation methods such as SMOTE-R have the potential to distort concept drifts and invalidate new augmented samples in time series problems \cite{moniz2017resampling, yosh2021TSBalancDist, eslami2022SMOTEDist}. This is because SMOTE-R can interpolate a new augmented sample using two samples observed at significantly different times, as long as they are close to both extreme values. If the distribution of the time series has drifted significantly between these times, then the synthesised samples do not preserve these systematic changes in distribution. Some strategies have been proposed to enable SMOTE-R to take into account the temporal dependency of time series data. A strategy called TS\_SMOTE \cite{martin2020DesignIssues} used \textit{dynamic time warping} (DTW) as an alternative way of interpolating samples. DTW uses the time stamps of the seed and neighbour pairs to generate synthetic time stamps for the SMOTE samples. It has successfully been used to estimate psychological and physiological states from thermal sensation and core body temperature time series \cite{yosh2021TSBalancDist}. Another strategy called C-SMOTE uses a variable size window to monitor concept drifts for online class imbalance learning \cite{bernardo2022CSMOTE}. Data indicating a potential concept drift gets saved into a separate window to ensure that C-SMOTE is always applied to data consistent with the current concept. 
 
Furthermore, SMOTE-R has been extended to handle concept drifts while solving extreme forecasting problems\cite{torgo2013smote}. The oversampling technique named SMOTE-R-bin partitions numeric time series data into bins of consecutive rare or common observations. This introduces temporal biases in the case selection process of SMOTE-R since a new sample can only be augmented from two samples existing within the same bin. These limitations preserve changes in distribution, as they ensure that interpolation can only take place between samples that are within the temporal vicinity of each other. Thus, it has the potential to combat the issue of concept drift.

\section{Methodology}
\subsection{Data embedding}

In our study, we focus on scaled univariate time series data in the form \([x_1,x_2,\dots, x_N]\), where \(N\) is the length of the time series and \(x_i \in [0,1]\) for   \(1\leq i \leq N\). We need to reconstruct the time series as a state-space vector in order to train the respective deep learning models for multistep-ahead prediction. This can be achieved using Taken's embedding theorem, which demonstrates that the delayed embedding reconstruction will retain all the important properties of the original time series data \cite{taken1981}. Therefore, given a time series \[X = [x_1,x_2,\dots, x_N]\] an embedded phase space can be constructed using sliding window of a fixed size ($D$) at regular interval ($T$) as \[X_t = [x_t,x_{t-T},\dots, x_{t-(D-1)T}]\]  where \(D\) is the embedding dimension and \(T\) is the time delay.  
\\
\\
Applying Taken's theorem, we define our inputs (features) with an embedding dimension (window size) \(D\) as  \[X_t = [x_t,x_{t-1},\dots, x_{t-(D-1)}].\] We define a multistep ahead prediction with \(P\) steps (prediction horizon) that will feature model outputs as, \[y_t = [x_{t+1},x_{t+2},\dots , x_{t+P}].\]
Hence, a deep learning model can use  \(X_t\)  to predict the next \(P\) steps given by \(y_t\).

\subsection{Relevance function}

Our study focuses on the development of models that can accurately predict extreme values in time series data when compared to conventional models. The extremeness of a data sample can be quantified using a relevance function and a relevance threshold, as demonstrated by Ribeiro \cite{ribeiro2011utility}. While there are other ways of measuring extremeness, a relevance function can be generalised to any extreme forecasting problem. Due to its wide applicability potential, it is the method we will use for our framework.
\\
\\
We define a relevance function \(\phi:\mathcal{X}\to [0,1]\), where \(\mathcal{X}\) is the time series data (samples). In this case,\(\phi\) maps an input time series sample to a relevance score between 0 and 1. The relevance scores that are closer to 1 indicate the sample is more extreme, while relevance scores closer to 0 indicate it is more common. Furthermore, we define a relevance threshold \(R_T \in [0,1]\), then
\begin{align*}
\text{extremes} &= \{x\in \mathcal{X} | \phi(x) \geq R_T\} \\
\text{commons} &= \{x\in \mathcal{X} | \phi(x) < R_T\}
\end{align*}

\begin{figure*}[ht!]
    \centering
    \includegraphics[scale=0.2]{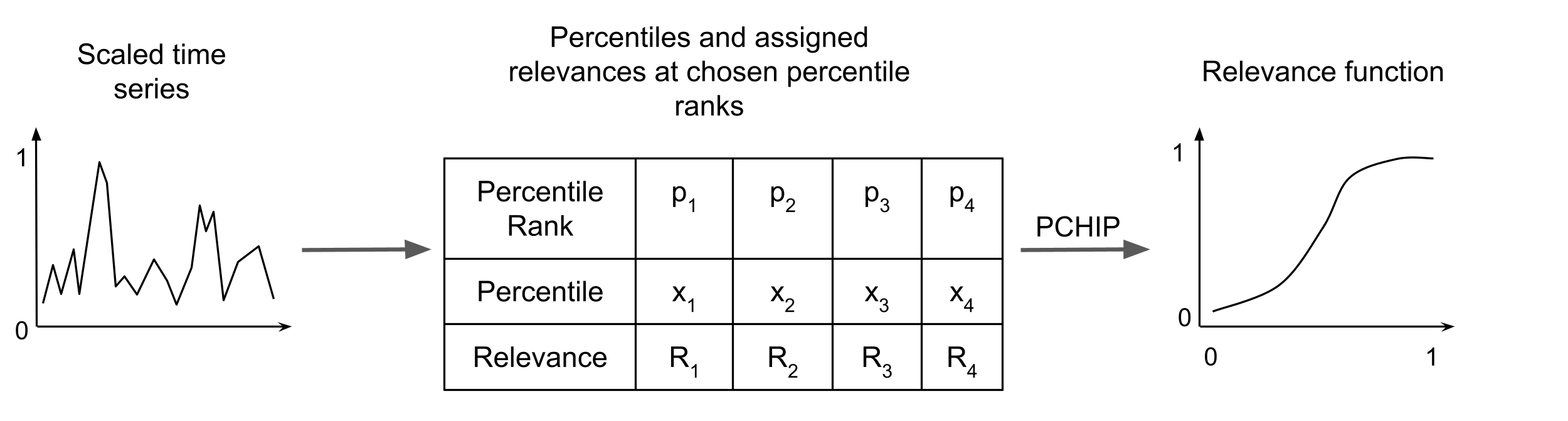}
    \caption{\textbf{Relevance function construction using PCHIP on percentiles}.}
    \label{fig:relevance function}
\end{figure*}

\noindent In real-world applications, the relevance function and threshold would be provided by field experts. In the absence of expert knowledge, there are several ways to construct a suitable relevance function. We utilise a \textit{piecewise cubic hermite interpolating polynomial} (PCHIP) constructed off of the boxplot statistics of a given data set, as proposed by Ribeiro \cite{ribeiro2011utility}. Specifically, we choose a set of percentile ranks, and compute the corresponding percentiles from the time series data. Then we attach a relevance score  for each percentile that will result in a set of percentile-relevance score pairs \(\{(x_k,R_k)\}\), where \(x_k\) is the scaled value of the time series and \(R_k\) is the associated relevance score. We apply  PCHIP  to this set of pairs to generate a relevance function, as displayed in Figure \ref{fig:relevance function}. 

\noindent Furthermore, since a relevance threshold specific to a dataset will be unavailable due to  the absence of expert knowledge, a range of thresholds will be used in our study for testing. We want to test a variety of relevance thresholds to ensure that the choice of relevance threshold does not impact the relative performance of the data augmentation techniques and deep learning models. There are three possible scenarios under which the extreme value forecasting problems will fall. The first case is when the extremes occur at both tails, i.e. both extremely large and small values will be classified as extreme. The other two cases are when there are either only large extremes or only small extremes. In these cases, we apply only an upper limit or a lower limit, respectively. The quantiles and their corresponding relevance scores should be picked judiciously to control which samples are considered as extreme.

\subsection{Relevance function  for  multistep-ahead prediction}
\label{sec:multistep}
Let \(y_t = [x_{t+1}, \dots , x_{t+P}]\) be a sliding window selected from a univariate time series data. We can define   relevance functions for a \(P\)-step window that includes maximum, minimum, average, and first step:
\begin{enumerate}
    \item Maximum 
    \begin{equation}
    \phi_{max}(y_t) = \max_{1\leq i \leq P} {\phi (x_{t+i}})
    \label{eq:maximum}
    \end{equation}

    \item Minimum \[\phi_{min}(y_t) = \min_{1\leq i \leq P} {\phi (x_{t+i}})\] \\
    \item Average \[\phi_{avg}(y_t) =  \frac{1}{P} \sum_{i=1}^P \phi(x_{t+i})\] \\
    \item First step \[\phi_{first}(y_t) = \phi(x_t)\]
\end{enumerate}

 In our study, we use the maximum relevance function since we want to predict an extreme value several time steps in advance.  Hence, as long as one of the time steps is relevant, the entire sample will be considered relevant (extreme). Note that a relevant sample refers to samples with a relevance score greater than the relevance threshold; in other words, a relevant sample is an extreme sample.
 
The original time series is embedded into input features \(X_t\) and output targets \(y_t\). Let \(\mathcal{\bar{X}}\) be the set of input features and \(\mathcal{Y}\) be the set of output targets. We use the maximum relevance function and  define the extreme and common samples using Equation \ref{eq:relevance}: 
\begin{align}
\text{extremes} &= \{(X_t,y_t)\in \mathcal{\bar{X}} \times \mathcal{Y} | \phi_{max}(y_t) \geq R_T\} \nonumber \\ 
\text{commons} &= \{(X_t,y_t)\in \mathcal{\bar{X}} \times \mathcal{Y} | \phi_{max}(y_t) < R_T\}  
\label{eq:relevance}
\end{align}

\subsection{Framework} 

\begin{figure*}[ht!]
    \centering
    \includegraphics[width=\textwidth]{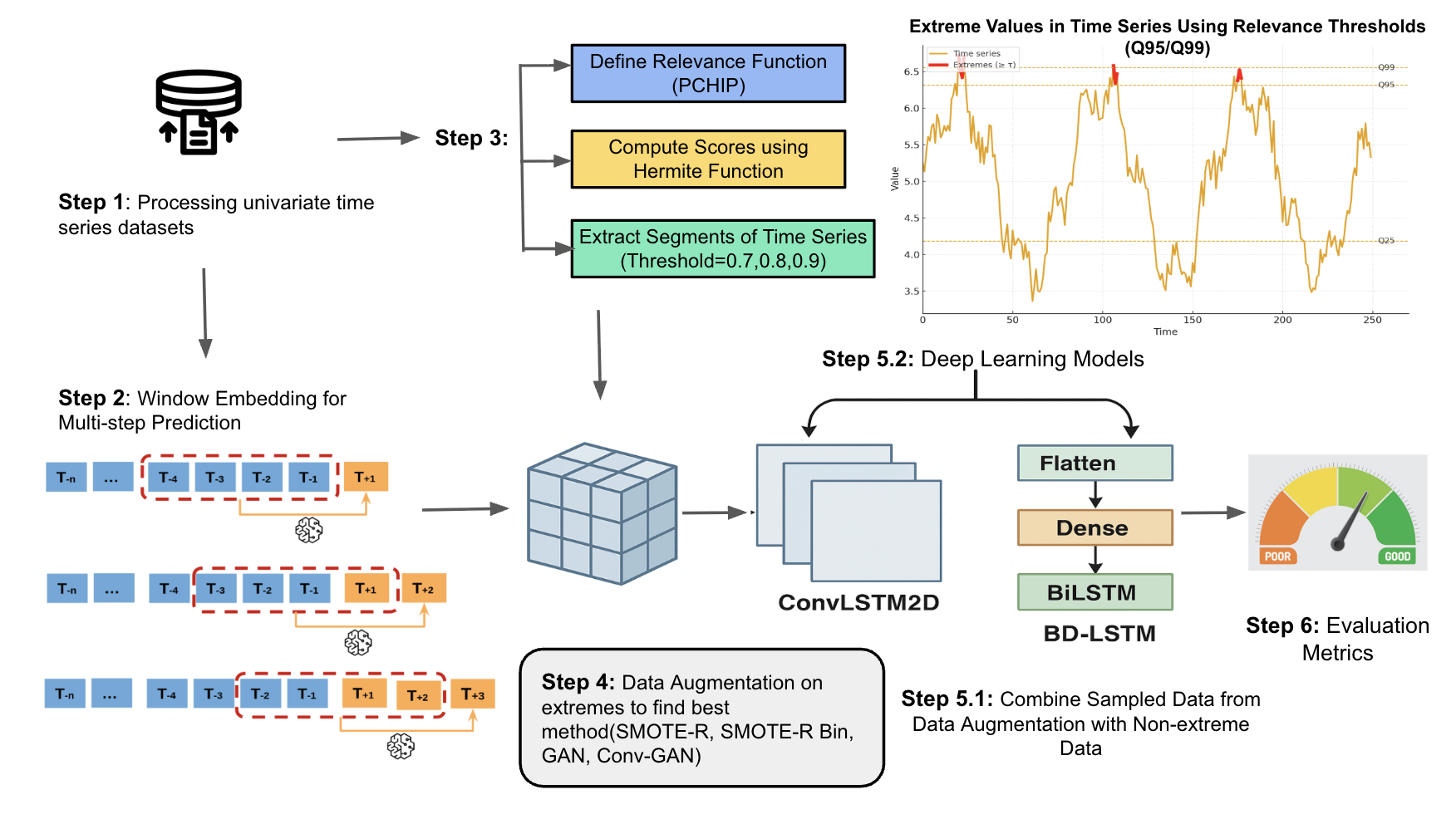}
    \caption{\textbf{Relevance-based framework for extreme value forecasting using data augmentation and deep learning.}}
    \label{fig:framework}
\end{figure*}

 Our relevance-based extreme value forecasting framework (Figure \ref{fig:framework}) is structured into sequential steps that collectively address the primary objectives of this study. 

We begin with processing univariate time series datasets, which are formulated as a multi-step ahead prediction problem (Step 1).  We scale the respective datasets and split ithem nto training and testing subsets. We reconstruct each time series into a state-space embedding through sliding windows (Step 2), following Taken’s theorem \cite{taken1981}, in order to prepare suitable input–output pairs for deep learning models. 

In Step 3, we apply the \text{relevance function} (PCHIP) and extract extreme values from the time series data. Due to the absence of expert-provided thresholds for extremes in these datasets, we construct generalised relevance functions using boxplot statistics for the scaled data, interpolated through a PCHIP function. We compute a maximum relevance score using a Hermite function for each sample (window). We then separate the samples into extreme and common classes based on selected relevance thresholds ($\tau \in \{0.7,0.8,0.9\}$). Evaluating multiple thresholds ensures that the framework remains robust and consistent across different choices of $\tau$. 

 We perform data augmentation on extremes (Step 4) by combining both traditional and generative methods to address the rarity of extreme samples. We apply SMOTE-R and SMOTE-R-bin together with 1D-GAN and 1D-Conv-GAN to enrich the representation of rare events. These approaches expand the pool of extreme samples and help reduce the imbalance between extreme and common values. We construct balanced training sets by combining the augmented extreme samples with non-extreme data. These datasets then serve as input to two baseline deep learning architectures (Step 5): a one-dimensional Convolutional LSTM (ConvLSTM2D) and a Bidirectional LSTM (BD-LSTM). Both models are designed to capture temporal dependencies in sequential data, and we systematically test them under different augmentation strategies to examine their comparative performance.

 Finally, to evaluate forecasting accuracy (Step 6), we adopt both conventional and relevance-based metrics. Alongside the standard RMSE, we calculate the Signal Extreme Ratio (SER) across percentiles ranging from 1\% to 75\%. This approach allows us to capture not only overall error but also the models’ sensitivity in predicting extreme events, providing a more comprehensive assessment.
 This stepwise framework enables systematic comparison of resampling strategies (e.g., SMOTE-R variants vs. GAN-based approaches) under multiple relevance thresholds. Moreover, the integration of relevance-based augmentation with deep learning models provides a structured approach for addressing the inherent imbalance of extremes in time series forecasting.

\subsection{Evaluation Metrics}


 The most commonly used evaluation metrics for regression (time series prediction/forecasting) problems include the Mean Squared Error (MSE), Root Mean Squared Error (RMSE), Mean Absolute Error (MAE), and Mean Absolute Percentage Error (MAPE)~\cite{botchkarev2018performance}. However, these evaluation metrics are not suitable for evaluating models within the relevance framework, since they are not ideal for extreme value forecasting \cite{ribeiro2008utility}. These metrics treat all observations with equal significance. In extreme value forecasting, we want to prioritise predicting the extreme values correctly. Additionally, there only exists a minute number of extreme samples compared to common samples in the datasets. Therefore, the error from the prediction of the commons will have a larger contribution to the evaluation metric than the prediction of extremes. Naturally, these metrics will favour models that predict common values accurately. For example, a forecasting model might appear to perform well according to the RMSE because it predicts the majority of common cases correctly while predicting the few extremes poorly. 
 
 Ribeiro and Moniz \cite{ribeiro2020imbalanced} proposed alternative metrics for extreme value forecasting, such as the Squared Error-Relevance (SER). On a dataset featuring extremes \(\mathcal{E}\), SER can be defined with respect to the relevance threshold \(R_T\) as follows: \[SER_{R_T} = \sum_{y_i \in \mathcal{E}} (y_i - \hat{y}_i)^2\], where \(y_i\) is an extreme sample and \(\hat{y}_i\) is the prediction of the corresponding sample. 




However, we are interested in forecasting both the common and extreme values with high accuracy. In our framework, we use RMSE and SER as our primary metrics and also evaluate the case-weight as a metric for extreme forecasting problems.



\subsection{Technical Details}
\label{sec:technical}
We compare data augmentation (resampling) methods on a variety of univariate time series data sets with various distributions. We used the \textit{Python} function \textit{dropna()} from the \textbf{Pandas} library to remove non-available (NA) observations. Following this, we used the  \textit{MinMaxScaler()} from the \textit{Sklearn} library to scale the observations to fit within the range of 0 to 1. This is a standard pre-processing technique commonly used in forecasting tasks because it helps forecasting models recognise patterns in time series and converge faster \cite{patro2015normalization}. We also attempted to use as many standardised libraries as possible to allow for easy replication and comparative testing in future work. For generating the PCHIP relevance function, we used the \textit{PchipInterpolator()} from the \textit{SciPy} package. We used Keras for all the deep learning models and PyTorch to implement GAN. 
 
Due to the computational power required to perform the iterative approach, a different device was used to leverage \textit{CUDA}. However, there were issues with applying a ReLU activation on the LSTM when using \textit{CUDA}. So, for the iterative approach, Tanh was used instead of ReLU, and so the iterative results should not be empirically compared against the other results.
 
\subsection{Data and Experiment setup}

 We conducted the experiments using five datasets, including both synthetic and real-world applications:
 \begin{enumerate}
     \item Bike:  contains bike-sharing records in London, sourced from the Kaggle  \cite{bikesharingdata}, which is common for evaluation of extreme value forecasting models. For example, Moniz et al. \cite{moniz2017resampling} used an SMOTE-R-based model using the bike count dataset.
     \item Lorenz: a synthetic dataset generated from the Lorenz attractor  \cite{lorenz1963deterministic}, which has become a benchmark for deterministic chaotic time series prediction.
     \item Sunspot: This dataset records historical sunspot counts, exhibiting long-term periodic variation. It has been extensively used in time series modelling, particularly suitable for validating the model's performance in forecasting relatively stationary yet nonlinearly trending sequences.
     \item Cyclone: We utilise datasets about cyclone wind-intensity from the  (South Pacific Ocean (SPO) and South Indian Ocean (SIO) \cite{cyclone} extracted from the Joint Typhoon Warning Centre (JTWC).  
 \end{enumerate}

Some of the datasets are multivariate but will be treated as univariate to predict a single variable (wind-intensity). Specifically, the Bike dataset will attempt to predict the \textit{count} variable, which is the number of bikes being shared. The Cyclone dataset will attempt to predict the \textit{wind intensity} variable, which is the wind intensity of the cyclones.  All data was scaled into the range [0,1] using a min-max-scaler. The data was embedded into windows by Taken's theorem with an embedding dimension (window size) of 5, and the output time horizon (number of steps) was also 5. A 70/30 training-test data split was used, with 70\% used for training and 30\% for testing. This was followed by an exploratory analysis of the data sets, which included the construction of a relevance function for each data set. 

 We use these resampling strategies to generate extremes of the form $(X_t, y_t)$ as defined in Equation~\ref{eq:relevance}. The resampling strategies produce both the input window and the output target for each extreme sample. We evaluate each resampling method using deep learning models, specifically ConvLSTM and BD-LSTM, trained on the resampled data. Since common values dominate the dataset, traditional metrics such as RMSE primarily reflect the error on common cases and may obscure model performance on rare extremes. For example, no-resampling may achieve a low RMSE, while performing poorly on the extreme samples. To address this issue, we adopt SER (Squared Error-Relevance) which focus on the model’s ability to predict extreme values.  

We summarise the architecture of the models used in this study in Table~\ref{tab:architecture}. During the data augmentation stage, we use GAN-based models, including 1D-GAN and 1D-Conv-GAN, to generate synthetic samples. These GANs are trained using the Adam optimiser~\cite{kingma2014adam} and employ binary cross-entropy as the loss function. We adopt deep learning models such as ConvLSTM and BD-LSTM for the forecasting stage using MSE loss function.

\begin{table*}[ht!]
    \centering
    \begin{tabular}{|c|c|c|}
        \hline
        Model & Hidden Layers & Details\\
        \hline
         1D-GAN generator& 3 fully connected & (fc\(_1\),fc\(_2\),fc\(_3\)) = (64,128,256) \\
         \hline
         1D-GAN discriminator& 3 fully connected & (fc\(_1\),fc\(_2\),fc\(_3\)) = (256,128,64) \\
         \hline
         1D-Conv-GAN& 2 convolutional& c\(_1\) = (filter = 256, kernel size = 3, \\
         generator&&     stride = 1, padding = 0) \\
         && c\(_2\) = (filter = 128, kernel size = 3) \\
         &&     stride = 1, padding = 0) \\
         & 1 fully connected& fc\(_1\) = (50) \\
         \hline
         1D-Conv-GAN& 2 transposed convolutional& tc\(_1\) = (filter = 128, kernel size = 3) \\
         discriminator&&     stride = 1, padding = 0) \\
         && tc\(_2\) = (filter = 256, kernel size = 3) \\
         &&     stride = 1, padding = 0) \\
         & 1 fully connected& fc\(_1\) = (50) \\
         \hline
         ConvLSTM&  1 ConvLSTM2D layer (filters = 64)& ConvLSTM2D: kernel size = (1,1)\\
 & Flatten, 1 Dense layer&Dense = N steps-ahead (output neurons)\\
         \hline
         BD-LSTM& 2 Bi-directional LSTM layers (units = hidden) & LSTM activation = ReLu\\
 & 1 Dense layer&Dense = N steps-ahead (output neurons)\\
        \hline
    \end{tabular}
    \caption{\textbf{Summary of architectures for forecasting models and GAN-based resampling strategies}}
    \label{tab:architecture}
\end{table*}

\section{Results}

 \subsection{Data exploration} 

We begin with an exploratory analysis to gain an intuitive understanding of how extremes are identified and characterised under the relevance-based framework. As this study involves five datasets with distinct features and distributions, we select the cyclone dataset as a representative example.  

\begin{figure*}[htbp!]
     \centering
     \begin{subfigure}[b]{1\textwidth}
         \centering
         \includegraphics[width=\textwidth]{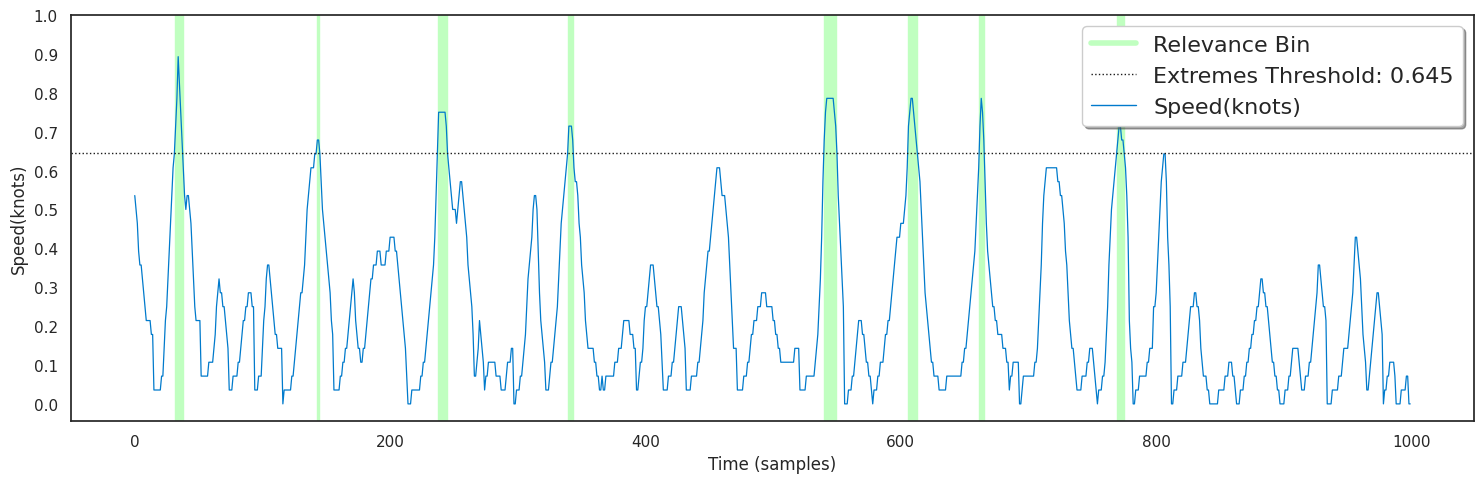}
         \caption{Cyclone time series with bins corresponding to a 0.7 relevance threshold}
         \label{fig:cyclone time series}
     \end{subfigure}
     \begin{subfigure}[b]{1\textwidth}
         \centering
         \includegraphics[width=\textwidth]{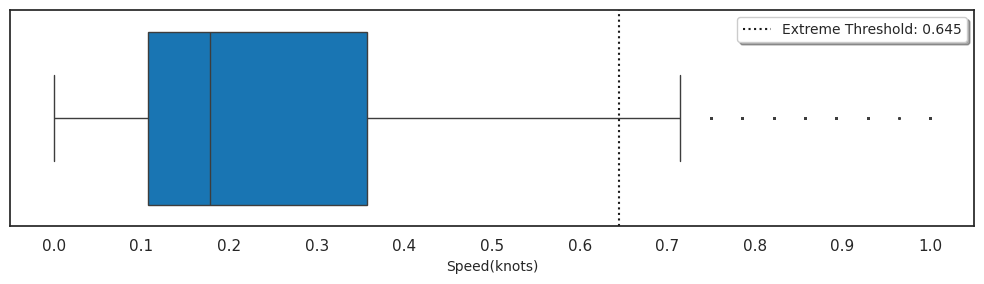}
         \caption{Cyclone boxplot}
         \label{fig:cyclone boxplot}
     \end{subfigure}
     \begin{subfigure}[b]{1\textwidth}
         \centering
         \includegraphics[width=\textwidth]{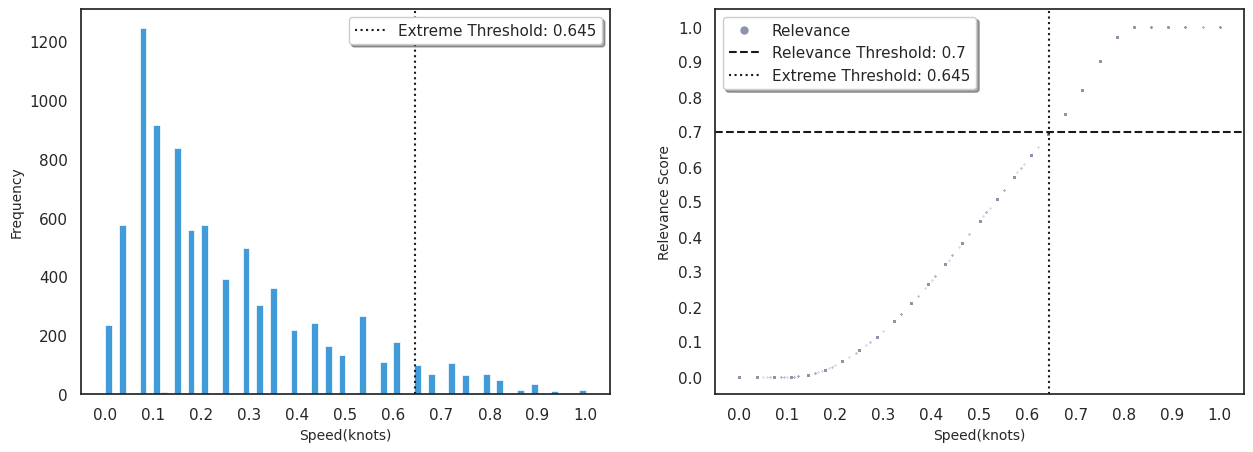}
         \caption{Distribution of Cyclone time series data as well as the distribution of the extremes at a 0.7 relevance threshold. Also shows the relevance function constructed using PCHIP and the conversion between relevance threshold and extreme threshold.}
         \label{fig:cyclone relevance}
     \end{subfigure}
        \caption{Cyclone dataset visualisation}
        \label{fig:cyclone}
\end{figure*}

Figure~\ref{fig:cyclone} illustrates the identification and distribution of extremes at a relevance threshold of 0.7. Panel (a) presents the time series with the extreme segments highlighted in green corresponding to consecutive time steps that exceed the threshold. This makes the sparsity and clustering of extreme events within the series immediately visible. Panel (b) shows the boxplot of the dataset, where the dashed line indicates the extreme threshold derived from the relevance function, situating the extremes within the overall distribution. Panel (c) further combines the histogram of the target distribution with the PCHIP-based relevance function, showing both the distributional characteristics of the data and the mapping to the extreme threshold. In this way, the framework not only delineates the boundary of extremes but also reveals how the proportion of extremes varies under different threshold settings.  

It is worth noting that at a relevance threshold of 0.7, the corresponding extreme threshold is 0.645. This observation suggests that the same relevance level may be assigned to different extreme thresholds across datasets. In other words, the definition of extremes is not fixed but is inherently dependent on the distribution of the data itself. For this reason, in the subsequent experiments, we pay particular attention to how relevance-based thresholding shapes the identification of extremes and how this, in turn, influences the performance of different resampling strategies and forecasting models.

\subsection{Baseline Resampling Strategies}

We evaluated the performance of four mainstream baseline resampling methods in the context of extreme value forecasting, namely SMOTER-regular, SMOTER-bin, 1D-GAN and 1D-Conv-GAN, and compared them with the no-resampling strategy. We run all experiments using the BD-LSTM model, and assess the model performance using two evaluation metrics: SER@5\%, and RMSE. We perform the evaluation using two representative datasets (Bike and Cyclone), under three different relevance thresholds  $\tau \in {0.7,0.8,0.9}$, to examine the stability and generalizability of the methods across varying definitions of extreme values. Table \ref{tab:resampling_performance} summarises the training and testing performance in all experimental configurations.  In the Bike dataset, 1D-Conv-GAN exhibits the best performance at $\tau$ = 0.7, achieving the lowest test SER (0.1101) while maintaining a competitive RMSE (0.0755), slightly outperforming SMOTER-regular, which records an SER of 0.1308 and RMSE of 0.0763. This indicates that Conv-GAN is effective in capturing extreme fluctuations under looser thresholds. However, as the relevance threshold increases to 0.8 and 0.9, the advantage clearly shifts to SMOTER-regular, which delivers the lowest RMSE values (0.1008 and 0.0968) with moderate SER levels, reflecting its robustness under stricter definitions of extremes. In contrast, SMOTER-bin and the GAN-based approaches deteriorate considerably as thresholds tighten, with 1D-GAN in particular showing unstable behaviour, reaching an SER of 0.3536 at $\tau$  = 0.9. These results suggest that Conv-GAN is useful for relatively lenient settings, but SMOTER-regular provides more reliable performance as extremes become rarer.

For the Cyclone datasets, SMOTE-based methods consistently outperform GAN-based approaches across thresholds, though their relative strengths vary. At $\tau$ = 0.7, SMOTER-bin achieves the lowest SER (0.0840), while SMOTER-regular attains the best RMSE (0.0832), indicating complementary advantages. At $\tau$ = 0.8, SMOTER-bin continues to yield the lowest SER (0.0853) but with higher RMSE (0.0989), whereas SMOTER-regular maintains a more balanced trade-off (SER = 0.1099, RMSE = 0.1087). Under the strictest threshold ($\tau$  = 0.9), SMOTER-regular achieves the most favourable performance, simultaneously recording the lowest SER (0.0745) and RMSE (0.0855). GAN-based methods, in contrast, degrade across all thresholds, reflecting instability and poor generalisation in volatile regimes. These findings highlight that SMOTER-bin is advantageous at moderate thresholds where extreme events are more frequent, while SMOTER-regular is the more robust choice as thresholds tighten and volatility increases.

In terms of data augmentation strategies, we both 1D-GAN and 1D-Conv-GAN exhibit high variability, with unstable SER and RMSE values that suggest potential training instability or architectural constraints. This weakness is most apparent in the Cyclone dataset under higher relevance thresholds, where GAN-based methods consistently lag behind SMOTE-based approaches, underscoring their limited robustness in extreme value forecasting. SMOTER-bin demonstrates greater adaptability to complex and volatile datasets, whereas SMOTER-regular achieves more reliable performance in relatively stationary series. Across both datasets, the resampling strategies consistently outperform the no-resampling baseline in terms of SER and RMSE, confirming the effectiveness of relevance-guided data augmentation in enhancing deep learning models for forecasting rare and extreme values.

\begin{table*}[ht!]
\begin{center}
\begin{tabular}{lclcccc}
    \hline
     Dataset&Relevance &Sampling Strategy&\multicolumn{2}{c}{SER-5\%}& \multicolumn{2}{c}{RMSE} \\
     &Threshold&&Train&Test&Train&Test\\
    \hline
     Bike&&no-resampling&0.1049&\cellcolor{red!40}0.1007&\cellcolor{orange!40}0.0621&\cellcolor{red!40}0.0668\\
 &0.7&SMOTER-regular&\cellcolor{orange!40}0.0623&0.1308&\cellcolor{red!40}0.0568&0.0763\\
 &&SMOTER-bin&0.1239&0.1815&0.0894&0.1152\\
 &&1D-GAN&0.0909&0.2145&0.0926&0.1093\\
 &&1D-Conv-GAN&\cellcolor{red!40}0.0590&\cellcolor{orange!40}0.1101&0.0665&\cellcolor{orange!40}0.0755\\
    \hline
     &0.8
    &SMOTER-regular&0.0866&\cellcolor{orange!40}0.1627&\cellcolor{orange!40}0.0785&\cellcolor{orange!40}0.1008\\
 &&SMOTER-bin&0.1498&0.2149&0.1195&0.1548\\
 &&1D-GAN&\cellcolor{orange!40}0.1039&0.2299&0.1176&0.1497\\
 &&1D-Conv-GAN&\cellcolor{red!40}0.0734&0.1784&0.0921&0.1097\\
    \hline
     &0.9
    &SMOTER-regular&\cellcolor{red!40}0.0782&\cellcolor{orange!40}0.1559&\cellcolor{orange!40}0.0749&\cellcolor{orange!40}0.0968\\
 &&SMOTER-bin&0.1933&0.2862&0.1318&0.1792\\
 &&1D-GAN&\cellcolor{orange!40}0.0978&0.3536&0.1409&0.1892\\
 &&1D-Conv-GAN&0.1387&0.3374&0.1208&0.1637\\
    \hline
 Cyclone (SPO)& & no-resampling& 0.1607& \cellcolor{orange!40}0.0854& \cellcolor{red!40}0.0865& \cellcolor{red!40}0.0696\\
     & 0.7& SMOTER-regular& \cellcolor{orange!40}0.1172& 0.0978& 0.0943& \cellcolor{orange!40}0.0832\\
 & & SMOTER-bin& \cellcolor{red!40}0.1151& \cellcolor{red!40}0.0840& 0.1033& 0.0876\\
 & & 1D-GAN& 0.1560& 0.1854& 0.1049& 0.1156\\
 & & 1D-Conv-GAN& 0.1335& 0.1121& \cellcolor{orange!40}0.0918& 0.0935\\
 \hline
 & 0.8& SMOTER-regular& \cellcolor{red!40}0.1039& 0.1099& 0.1100& 0.1087\\
     & & SMOTER-bin& \cellcolor{orange!40}0.1220& \cellcolor{red!40}0.0853& 0.1111& \cellcolor{orange!40}0.0989\\
 & & 1D-GAN& 0.1239& 0.1487& 0.1050& 0.1163\\
 & & 1D-Conv-GAN& 0.1596& 0.1415& \cellcolor{orange!40}0.1038& 0.1128\\
 \hline
 & 0.9& SMOTER-regular& \cellcolor{red!40}0.0895& \cellcolor{red!40}0.0745& \cellcolor{orange!40}0.0999& \cellcolor{orange!40}0.0855\\
 & & SMOTER-bin& \cellcolor{orange!40}0.0991& 0.1032& 0.1164& 0.1066\\
 & & 1D-GAN& 0.1639& 0.1567& 0.1219& 0.1393\\
 & & 1D-Conv-GAN& 0.1164& 0.1024& 0.0928& 0.1082\\
 \hline
\end{tabular}
\end{center}
\label{bike lstm}
\caption{ \textbf{Performance Comparison of Resampling Strategies Across Relevance Thresholds using }\textbf{BD-LSTM model.} For each relevance threshold there is highlighting: \textcolor{red}{red} indicates the best performing strategy for the metric, \textcolor{orange}{orange} indicates second best strategy.}
\label{tab:resampling_performance}
\end{table*}

\subsection{Performance over multiple time steps}

\begin{figure*}[H]
     \centering
     
     \begin{subfigure}[b]{0.48\textwidth}
         \centering
         \includegraphics[width=\textwidth]{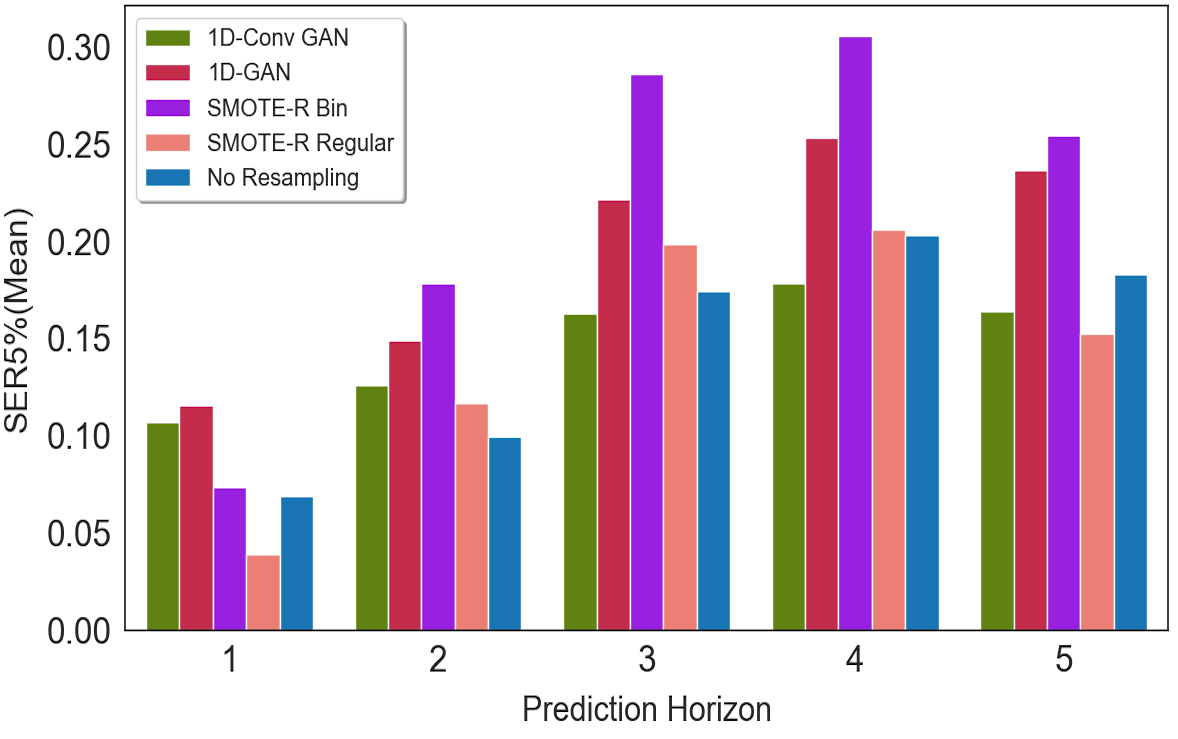}
         \caption{Bike Dataset}
         \label{fig:three sin x}
     \end{subfigure}
      \begin{subfigure}[b]{0.48\textwidth}
         \centering
         \includegraphics[width=\textwidth]{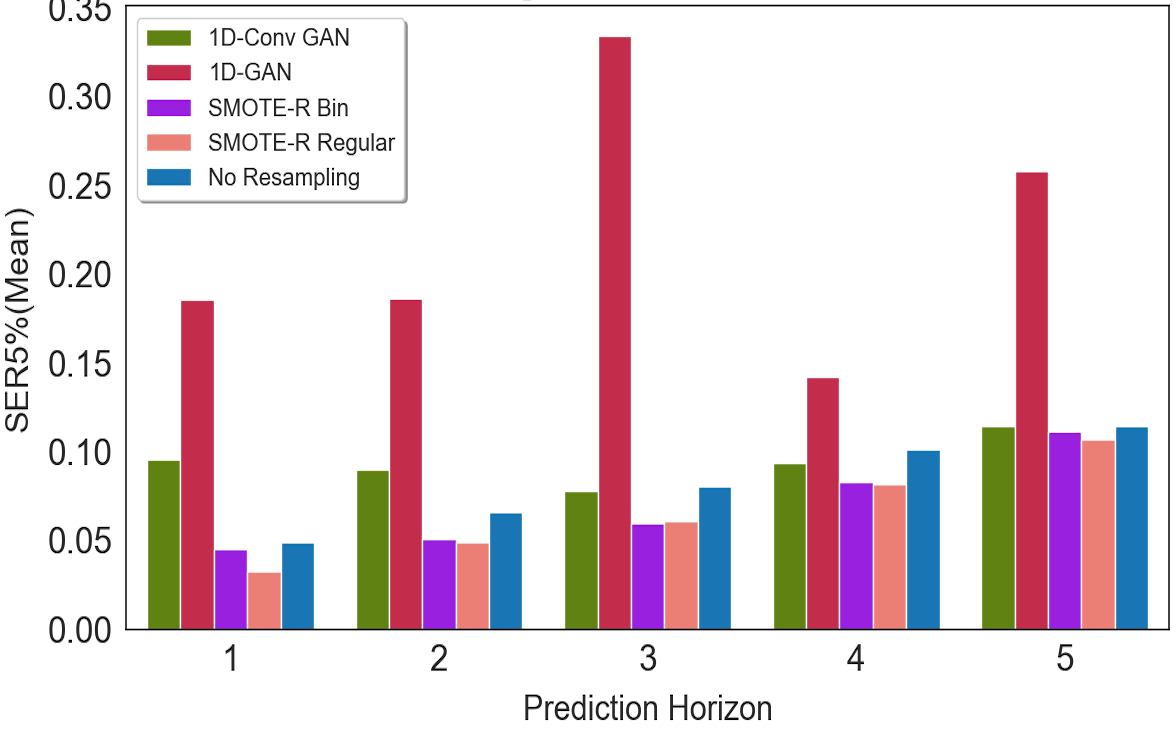}
         \caption{Cyclone Dataset}
         \label{fig:three sin x}
     \end{subfigure}
        \caption{\textbf{Performance of Resampling Strategies on 5-Step Ahead SER-5\% for the Bike and Cyclone Datasets Using BD-LSTM}}
        \label{fig:5 steps}
\end{figure*}

We examine the SER-5\% across five prediction horizons using the BD-LSTM model to evaluate the effectiveness of different resampling strategies in multi-step extreme forecasting, as shown in Figure ~\ref{fig:5 steps} for both the Cyclone and Bike datasets. Overall, errors increase with longer horizons, reflecting the growing difficulty of capturing extremes further into the future. However, the relative performance of individual strategies varies across datasets and prediction steps. In the Bike dataset, SER-5\% highlights clear differences among resampling strategies. In the first prediction step, SMOTER-bin and SMOTER-regular achieve the lowest errors, indicating strong short-term performance. However, from the second step onward, the error of SMOTER-bin increases sharply, making it less effective for longer horizons. By contrast, SMOTER-regular maintains stable performance across steps, demonstrating robustness. By the fifth step, both 1D-Conv-GAN and SMOTER-regular emerge as the most competitive strategies, suggesting their suitability for long-horizon extreme forecasting.

In the Cyclone dataset, SMOTER-regular achieves the lowest SER-5\% at the first step and maintains strong performance thereafter. From the second step onward, its results converge with those of SMOTER-bin, with both strategies consistently outperforming other methods. This suggests that SMOTE-based approaches are effective in volatile but structured series, where extremes occur with some regularity. In contrast, 1D-GAN performs the worst across all steps, with steep error increases at longer horizons, indicating limited generalisation capacity. These results indicate that SMOTER-regular provides consistently strong performance across both datasets, especially in longer forecasting horizons. SMOTER-bin is competitive in structured series such as Cyclone, but less reliable in more variable settings like Bike, highlighting the importance of aligning resampling strategies with the underlying data characteristics in multi-step extreme forecasting.

\subsection{Deep Learning models using selected Resampling Strategies}

We now compare the two best-performing strategies, including SMOTER-bin and SMOTER-R, across all five datasets and both Conv-LSTM and BD-LSTM-based models. In this way, we verify whether the previously identified advantages of these strategies hold consistently across different data and model conditions.

 We first evaluate the performance of all model–resampling combinations across five datasets under varying extreme value thresholds, ranging from 1\% to 75\% SER. This validates the effectiveness and robustness of the proposed modelling and resampling strategies in capturing rare but critical outcomes. Table \ref{tab:DL table} reports the SER (mean and standard deviation (+/-)) for each configuration, averaged across ten runs. At the SER1\% level, BD-LSTM with SMOTER-regular achieves the lowest mean error on the Lorenz dataset (0.0117 ± 0.0014), while Conv-LSTM with SMOTER-bin performs best on Sunspot and South Pacific cyclone datasets (0.0206 ± 0.0021 and 0.0156 ± 0.0014, respectively). In contrast, the Bike dataset shows stronger results for no-resampling strategies, with BD-LSTM yielding 0.0214 ± 0.0012, indicating that synthetic resampling may amplify noise in high-variance datasets. Across different SER thresholds, SMOTER-bin generally maintains lower variance, especially on datasets such as  Sunspot, suggesting its robustness in controlling error.

 Table \ref{tab:DL table}  also reports that several combinations consistently underperform; for example, BD-LSTM with no-resampling on Sunspot results in a notably high SER1\% of 0.1481 ± 0.0144, while SMOTER-bin with BD-LSTM reaches 0.1845 ± 0.0190 for the Bike dataset. These large variances suggest that some strategies may not generalise well under severe data imbalance or temporal irregularity. Moreover, configurations such as SMOTER-bin on Bike and Cyclone-SI exhibit high errors and variability, reflecting unstable model behaviour on noisy or low-frequency data.  These findings further support the role of relevance-based partitioning in aligning the sampling strategy with data irregularity, and suggest that ensemble learning can smooth out performance fluctuations caused by local overfitting. Across datasets, the results underscore the need to adapt resampling strategies to the temporal structure and distribution of extremes. 

\begin{table*}[htbp!]
\begin{center}
\resizebox{\textwidth}{!}{%
    \begin{tabular}{cccccccccll}\toprule
           dataset&  model&  sampling  strategy&  SER1\%&  SER2\%&  SER5\%&  SER10\%&  SER25\%&  SER50\%& SER75\%&RMSE\\\midrule
 & & & & & & & & & &\\
           Lorenz&  Conv-LSTM&  no-resampling&  \cellcolor{cyan!40}\makecell{0.0296 \\ $\pm$ 0.0352}&  \cellcolor{cyan!40}\makecell{0.0285 \\ $\pm$ 0.0338}&  \cellcolor{cyan!40}\makecell{0.0271 \\ $\pm$ 0.0324}&  \makecell{0.0264 \\ $\pm$ 0.0318}&  \makecell{0.0273 \\ $\pm$ 0.0340}&  \makecell{0.0218 \\ $\pm$ 0.0260}&\makecell{0.0211 \\ $\pm$ 0.0234}&\makecell{0.0214 \\ $\pm$ 0.0245}
\\
 & & SMOTER-regular& \cellcolor{orange!40}\makecell{0.0186 \\ $\pm$ 0.0314}& \cellcolor{orange!40}\makecell{0.0186 \\ $\pm$ 0.0319}& \cellcolor{orange!40}\makecell{0.0190 \\ $\pm$ 0.0326}& \cellcolor{orange!40}\makecell{0.0201 \\ $\pm$ 0.0335}& \cellcolor{orange!40}\makecell{0.0230 \\ $\pm$ 0.0365}& \cellcolor{orange!40}\makecell{0.0206 \\ $\pm$ 0.0313}& \cellcolor{orange!40}\makecell{0.0235 \\ $\pm$ 0.0321}&\makecell{0.0254 \\ $\pm$ 0.0361}\\
 & & SMOTER-bin& \makecell{0.0242 \\ $\pm$ 0.0322}& \makecell{0.0245 \\ $\pm$ 0.0328}& \makecell{0.0254 \\ $\pm$ 0.0321}& \cellcolor{cyan!40}\makecell{0.0270 \\ $\pm$ 0.0321}& \cellcolor{cyan!40}\makecell{0.0302 \\ $\pm$ 0.0342}& \cellcolor{cyan!40}\makecell{0.0286 \\ $\pm$ 0.0291}& \cellcolor{cyan!40}\makecell{0.0298 \\ $\pm$ 0.0297}&\cellcolor{cyan!40}\makecell{0.0326 \\ $\pm$ 0.0328}\\
 & BD-LSTM& no-resampling& \makecell{0.0278 \\ $\pm$ 0.0286}& \makecell{0.0269 \\ $\pm$ 0.0274}& \makecell{0.0262 \\ $\pm$ 0.0263}& \makecell{0.0265 \\ $\pm$ 0.0268}& \makecell{0.0288 \\ $\pm$ 0.0292}& \makecell{0.0247 \\ $\pm$ 0.0223}& \makecell{0.0240 \\ $\pm$ 0.0210}&\cellcolor{orange!40}\makecell{0.0233 \\ $\pm$ 0.0203}\\
 & & SMOTER-regular& \cellcolor{red!40}\makecell{0.0117 \\ $\pm$ 0.0207}& \cellcolor{red!40}\makecell{0.0124 \\ $\pm$ 0.0221}& \cellcolor{red!40}\makecell{0.0140 \\ $\pm$ 0.0247}& \cellcolor{red!40}\makecell{0.0158 \\ $\pm$ 0.0274}& \cellcolor{red!40}\makecell{0.0202 \\ $\pm$ 0.0319}& \cellcolor{red!40}\makecell{0.0203 \\ $\pm$ 0.0299}& \makecell{0.0253 \\ $\pm$ 0.0319}&\makecell{0.0274 \\ $\pm$ 0.0364}\\
 & & SMOTER-bin& \makecell{0.0216 \\ $\pm$ 0.0168}& \makecell{0.0216 \\ $\pm$ 0.0187}& \makecell{0.0227 \\ $\pm$ 0.0208}& \makecell{0.0247 \\ $\pm$ 0.0234}& \makecell{0.0283 \\ $\pm$ 0.0282}& \makecell{0.0272 \\ $\pm$ 0.0251}& \makecell{0.0285 \\ $\pm$ 0.0263}&\makecell{0.0310 \\ $\pm$ 0.0300}\\
 & & & & & & & & & &\\
 Cyclone-SPO& Conv-LSTM& no-resampling& \makecell{0.0783 \\ $\pm$ 0.0278}& \makecell{0.0843 \\ $\pm$ 0.0283}& \cellcolor{orange!40}\makecell{0.0813 \\ $\pm$ 0.0283}& \cellcolor{red!40}\makecell{0.0837 \\ $\pm$ 0.0250}& \cellcolor{red!40}\makecell{0.0836 \\ $\pm$ 0.0224}& \cellcolor{red!40}\makecell{0.0822 \\ $\pm$ 0.0205}& \cellcolor{red!40}\makecell{0.0797 \\ $\pm$ 0.0180}&\cellcolor{red!40}\makecell{0.0712 \\ $\pm$ 0.0164}\\
 & & SMOTER-regular& \cellcolor{orange!40}\makecell{0.0732 \\ $\pm$ 0.0105}& \makecell{0.0790 \\ $\pm$ 0.0116}& \makecell{0.0987 \\ $\pm$ 0.0187}& \makecell{0.1011 \\ $\pm$ 0.0103}& \makecell{0.0930 \\ $\pm$ 0.0204}& \makecell{0.0937 \\ $\pm$ 0.0423}& \makecell{0.0947 \\ $\pm$ 0.0540}&\makecell{0.0890 \\ $\pm$ 0.0590}
\\
 & & SMOTER-bin& \cellcolor{red!40}\makecell{0.0626 \\ $\pm$ 0.0109}& \cellcolor{red!40}\makecell{0.0692 \\ $\pm$ 0.0112}& \cellcolor{red!40}\makecell{0.0749 \\ $\pm$ 0.0123}& \cellcolor{orange!40}\makecell{0.0842 \\ $\pm$ 0.0074}& \makecell{0.0857 \\ $\pm$ 0.0111}& \makecell{0.0876 \\ $\pm$ 0.0204}& \makecell{0.0875 \\ $\pm$ 0.0268}&\makecell{0.0810 \\ $\pm$ 0.0297}
\\
 & BD-LSTM& no-resampling& \cellcolor{cyan!40}\makecell{0.0985 \\ $\pm$ 0.0490}& \cellcolor{cyan!40}\makecell{0.0934 \\ $\pm$ 0.0312}& \makecell{0.0892 \\ $\pm$ 0.0173}& \makecell{0.0892 \\ $\pm$ 0.0099}& \cellcolor{orange!40}\makecell{0.0844 \\ $\pm$ 0.0075}& \cellcolor{orange!40}\makecell{0.0830 \\ $\pm$ 0.0087}& \cellcolor{orange!40}\makecell{0.0802 \\ $\pm$ 0.0085}&\cellcolor{orange!40}\makecell{0.0718 \\ $\pm$ 0.0081}
\\
 & & SMOTER-regular& \makecell{0.0822 \\ $\pm$ 0.0144}& \makecell{0.0877 \\ $\pm$ 0.0137}& \cellcolor{cyan!40}\makecell{0.1048 \\ $\pm$ 0.0147}& \cellcolor{cyan!40}\makecell{0.1079 \\ $\pm$ 0.0095}& \cellcolor{cyan!40}\makecell{0.1001 \\ $\pm$ 0.0077}& \cellcolor{cyan!40}\makecell{0.0994 \\ $\pm$ 0.0256}& \makecell{0.0979 \\ $\pm$ 0.0382}&\makecell{0.0903 \\ $\pm$ 0.0431}
\\
 & & SMOTER-bin& \makecell{0.0744 \\ $\pm$ 0.0368}& \cellcolor{orange!40}\makecell{0.0765 \\ $\pm$ 0.0264}& \makecell{0.0839 \\ $\pm$ 0.0120}& \makecell{0.0925 \\ $\pm$ 0.0067}& \makecell{0.0968 \\ $\pm$ 0.0149}& \makecell{0.0993 \\ $\pm$ 0.0319}& \cellcolor{cyan!40}\makecell{0.0981 \\ $\pm$ 0.0411}&\cellcolor{cyan!40}\makecell{0.0903 \\ $\pm$ 0.0445}\\
 & & & & & & & & & &\\
 Bike& Conv-LSTM& no-resampling& \cellcolor{orange!40}\makecell{0.0772 \\ $\pm$ 0.0222}& \cellcolor{orange!40}\makecell{0.0829 \\ $\pm$ 0.0179}& \cellcolor{red!40}\makecell{0.0991 \\ $\pm$ 0.0091}& \cellcolor{orange!40}\makecell{0.1168 \\ $\pm$ 0.0052}& \cellcolor{orange!40}\makecell{0.1136 \\ $\pm$ 0.0113}& \cellcolor{orange!40}\makecell{0.1012 \\ $\pm$ 0.0174}& \cellcolor{orange!40}\makecell{0.0928 \\ $\pm$ 0.0210}&\cellcolor{orange!40}\makecell{0.0803 \\ $\pm$ 0.0213}
\\
 & & SMOTER-regular& \makecell{0.1041 \\ $\pm$ 0.0665}& \makecell{0.1027 \\ $\pm$ 0.0618}& \makecell{0.1043 \\ $\pm$ 0.0522}& \makecell{0.1250 \\ $\pm$ 0.0496}& \makecell{0.1362 \\ $\pm$ 0.0651}& \makecell{0.1180 \\ $\pm$ 0.0539}& \makecell{0.1042 \\ $\pm$ 0.0468}&\makecell{0.0909 \\ $\pm$ 0.0414}
\\
 & & SMOTER-bin& \makecell{0.1176 \\ $\pm$ 0.0476}& \makecell{0.1263 \\ $\pm$ 0.0478}& \makecell{0.1462 \\ $\pm$ 0.0400}& \makecell{0.1543 \\ $\pm$ 0.0336}& \cellcolor{cyan!40}\makecell{0.1563 \\ $\pm$ 0.0356}& \cellcolor{cyan!40}\makecell{0.1454 \\ $\pm$ 0.0364}& \cellcolor{cyan!40}\makecell{0.1342 \\ $\pm$ 0.0326}&\cellcolor{cyan!40}\makecell{0.1215 \\ $\pm$ 0.0291}
\\
 & BD-LSTM& no-resampling& \cellcolor{red!40}\makecell{0.0772 \\ $\pm$ 0.0214}& \cellcolor{red!40}\makecell{0.0815 \\ $\pm$ 0.0142}& \cellcolor{orange!40}\makecell{0.1033 \\ $\pm$ 0.0117}& \cellcolor{red!40}\makecell{0.1155 \\ $\pm$ 0.0107}& \cellcolor{red!40}\makecell{0.1071 \\ $\pm$ 0.0137}& \cellcolor{red!40}\makecell{0.0946 \\ $\pm$ 0.0185}& \cellcolor{red!40}\makecell{0.0853 \\ $\pm$ 0.0207}&\cellcolor{red!40}\makecell{0.0736 \\ $\pm$ 0.0196}\\
 & & SMOTER-regular& \makecell{0.1567 \\ $\pm$ 0.0484}& \makecell{0.1496 \\ $\pm$ 0.0456}& \makecell{0.1397 \\ $\pm$ 0.0395}& \makecell{0.1425 \\ $\pm$ 0.0382}& \makecell{0.1326 \\ $\pm$ 0.0398}& \makecell{0.1128 \\ $\pm$ 0.0382}& \makecell{0.0992 \\ $\pm$ 0.0354}&\makecell{0.0873 \\ $\pm$ 0.0353}
\\
 & & SMOTER-bin& \cellcolor{cyan!40}\makecell{0.1845 \\ $\pm$ 0.0328}& \cellcolor{cyan!40}\makecell{0.1900 \\ $\pm$ 0.0262}& \cellcolor{cyan!40}\makecell{0.1816 \\ $\pm$ 0.0200}& \cellcolor{cyan!40}\makecell{0.1702 \\ $\pm$ 0.0177}& \makecell{0.1542 \\ $\pm$ 0.0166}& \makecell{0.1374 \\ $\pm$ 0.0234}& \makecell{0.1240 \\ $\pm$ 0.0256}&\makecell{0.1125 \\ $\pm$ 0.0256}\\
 & & & & & & & & & &\\
 Sunspot& Conv-LSTM& no-resampling& \makecell{0.0986 \\ $\pm$ 0.1275}& \makecell{0.0958 \\ $\pm$ 0.1167}& \makecell{0.0940 \\ $\pm$ 0.1124}& \makecell{0.0934 \\ $\pm$ 0.1096}& \makecell{0.0861 \\ $\pm$ 0.1047}& \makecell{0.0757 \\ $\pm$ 0.0891}& \makecell{0.0681 \\ $\pm$ 0.0762}&\makecell{0.0597 \\ $\pm$ 0.0706}
\\
 & & SMOTER-regular& \makecell{0.0740 \\ $\pm$ 0.0331}& \makecell{0.0740 \\ $\pm$ 0.0331}& \makecell{0.0740 \\ $\pm$ 0.0331}& \makecell{0.0710 \\ $\pm$ 0.0301}& \makecell{0.0662 \\ $\pm$ 0.0284}& \makecell{0.0678 \\ $\pm$ 0.0329}& \makecell{0.0663 \\ $\pm$ 0.0372}&
\makecell{0.0618 \\ $\pm$ 0.0394}\\
 & & SMOTER-bin& \cellcolor{red!40}\makecell{0.0551 \\ $\pm$ 0.0206}& \cellcolor{red!40}\makecell{0.0551 \\ $\pm$ 0.0206}& \cellcolor{red!40}\makecell{0.0551 \\ $\pm$ 0.0206}& \cellcolor{red!40}\makecell{0.0551 \\ $\pm$ 0.0206}& \cellcolor{red!40}\makecell{0.0588 \\ $\pm$ 0.0216}& \cellcolor{orange!40}\makecell{0.0582 \\ $\pm$ 0.0248}& \cellcolor{orange!40}\makecell{0.0562 \\ $\pm$ 0.0302}&\cellcolor{orange!40}\makecell{0.0529 \\ $\pm$ 0.0335}
\\
 & BD-LSTM& no-resampling& \cellcolor{cyan!40}\makecell{0.1481 \\ $\pm$ 0.1253}& \cellcolor{cyan!40}\makecell{0.1467 \\ $\pm$ 0.1217}& \cellcolor{cyan!40}\makecell{0.1457 \\ $\pm$ 0.1191}& \cellcolor{cyan!40}\makecell{0.1449 \\ $\pm$ 0.1177}& \cellcolor{cyan!40}\makecell{0.1140 \\ $\pm$ 0.0842}& \cellcolor{cyan!40}\makecell{0.0920 \\ $\pm$ 0.0650}& \cellcolor{cyan!40}\makecell{0.0804 \\ $\pm$ 0.0548}&\cellcolor{cyan!40}\makecell{0.0691 \\ $\pm$ 0.0526}
\\
 & & SMOTER-regular& \makecell{0.1103 \\ $\pm$ 0.0971}& \makecell{0.1103 \\ $\pm$ 0.0971}& \makecell{0.1103 \\ $\pm$ 0.0971}& \makecell{0.1048 \\ $\pm$ 0.0962}& \makecell{0.0829 \\ $\pm$ 0.0610}& \makecell{0.0776 \\ $\pm$ 0.0437}& \makecell{0.0721 \\ $\pm$ 0.0389}&
\makecell{0.0650 \\ $\pm$ 0.0365}\\
 & & SMOTER-bin& \cellcolor{orange!40}\makecell{0.0611 \\ $\pm$ 0.0240}& \cellcolor{orange!40}\makecell{0.0611 \\ $\pm$ 0.0240}& \cellcolor{orange!40}\makecell{0.0611 \\ $\pm$ 0.0240}& \cellcolor{orange!40}\makecell{0.0601 \\ $\pm$ 0.0221}& \cellcolor{orange!40}\makecell{0.0591 \\ $\pm$ 0.0203}& \cellcolor{red!40}\makecell{0.0582 \\ $\pm$ 0.0224}& \cellcolor{red!40}\makecell{0.0561 \\ $\pm$ 0.0249}&\cellcolor{red!40}\makecell{0.0525 \\ $\pm$ 0.0270}\\
 & & & & & & & & & &\\
 Cyclone-SIO& Conv-LSTM& no-resampling& \cellcolor{red!40}\makecell{0.0571 \\ $\pm$ 0.0012}& \cellcolor{red!40}\makecell{0.0576 \\ $\pm$ 0.0014}& \cellcolor{red!40}\makecell{0.0608 \\ $\pm$ 0.0012}& \cellcolor{red!40}\makecell{0.0607 \\ $\pm$ 0.0006}& \cellcolor{red!40}\makecell{0.0661 \\ $\pm$ 0.0007}& \cellcolor{orange!40}\makecell{0.0697 \\ $\pm$ 0.0008}& \cellcolor{orange!40}\makecell{0.0678 \\ $\pm$ 0.0009}&\cellcolor{orange!40}\makecell{0.0605 \\ $\pm$ 0.0008}
\\
 & & SMOTER-regular& \makecell{0.0695 \\ $\pm$ 0.0035}& \makecell{0.0692 \\ $\pm$ 0.0035}& \makecell{0.0696 \\ $\pm$ 0.0022}& \makecell{0.0686 \\ $\pm$ 0.0020}& \makecell{0.0784 \\ $\pm$ 0.0019}& \makecell{0.0885 \\ $\pm$ 0.0038}& \makecell{0.0890 \\ $\pm$ 0.0045}&\makecell{0.0862 \\ $\pm$ 0.0039}
\\
 & & SMOTER-bin& \makecell{0.0576 \\ $\pm$ 0.0044}& \cellcolor{orange!40}\makecell{0.0576 \\ $\pm$ 0.0051}& \makecell{0.0655 \\ $\pm$ 0.0024}& \makecell{0.0665 \\ $\pm$ 0.0010}& \makecell{0.0779 \\ $\pm$ 0.0009}& \makecell{0.0864 \\ $\pm$ 0.0023}& \makecell{0.0872 \\ $\pm$ 0.0033}&\makecell{0.0838 \\ $\pm$ 0.0036}
\\
 & BD-LSTM& no-resampling& \cellcolor{orange!40}\makecell{0.0571 \\ $\pm$ 0.0033}& \makecell{0.0583 \\ $\pm$ 0.0039}& \cellcolor{orange!40}\makecell{0.0626 \\ $\pm$ 0.0034}& \cellcolor{orange!40}\makecell{0.0628 \\ $\pm$ 0.0022}& \cellcolor{orange!40}\makecell{0.0672 \\ $\pm$ 0.0019}& \cellcolor{red!40}\makecell{0.0696 \\ $\pm$ 0.0012}& \cellcolor{red!40}\makecell{0.0675 \\ $\pm$ 0.0011}&\cellcolor{red!40}\makecell{0.0601 \\ $\pm$ 0.0010}\\
 & & SMOTER-regular& \cellcolor{cyan!40}\makecell{0.0810 \\ $\pm$ 0.0166}& \cellcolor{cyan!40}\makecell{0.0812 \\ $\pm$ 0.0162}& \cellcolor{cyan!40}\makecell{0.0898 \\ $\pm$ 0.0137}& \cellcolor{cyan!40}\makecell{0.0906 \\ $\pm$ 0.0143}& \cellcolor{cyan!40}\makecell{0.0991 \\ $\pm$ 0.0171}& \cellcolor{cyan!40}\makecell{0.1078 \\ $\pm$ 0.0373}& \cellcolor{cyan!40}\makecell{0.1099 \\ $\pm$ 0.0525}&\cellcolor{cyan!40}\makecell{0.1059 \\ $\pm$ 0.0545}
\\
 & & SMOTER-bin& \makecell{0.0677 \\ $\pm$ 0.0199}& \makecell{0.0683 \\ $\pm$ 0.0198}& \makecell{0.0719 \\ $\pm$ 0.0107}& \makecell{0.0727 \\ $\pm$ 0.0067}& \makecell{0.0862 \\ $\pm$ 0.0041}& \makecell{0.0965 \\ $\pm$ 0.0099}& \makecell{0.0927 \\ $\pm$ 0.0100}&\makecell{0.0853 \\ $\pm$ 0.0093}\\
    \end{tabular}
}
\end{center}
\caption{Deep learning model performance (SER) using different resampling strategies. For each relevance threshold, red indicates the best performing strategy for the metric,  orange indicates the second best strategy, and blue indicates the worst performing strategy. The cyclones are categorised by the South Indian Ocean (Cyclone-SIO) and the South Pacific Ocean (Cyclone-SPO).}
\label{tab:DL table}
\end{table*}

 Table \ref{tab:summaryprediction} examines how the model strategy combinations (deep learning model and data augmentation (resampling strategy) perform across datasets with distinct temporal characteristics. We find that BD-LSTM performs the best in the case of Lorenz and Bike, while Conv-LSTM consistently outperforms others on the remaining datasets. SMOTER-regular achieves the lowest error on Lorenz, whereas SMOTER-bin performs well on periodic datasets such as Sunspot and Cyclone-SPO, likely due to its ability to enrich extreme samples. In contrast, strategies such as no-resampling underperform on datasets with limited extremes. These results emphasise that no single resampling strategy excels universally; rather, performance hinges on the interaction between temporal patterns and class imbalance, highlighting the need for context-aware design.

\begin{table*}[htbp!]
\begin{center}
    \begin{tabular}{ccccl}\toprule
         Dataset&  Best Model&  Best strategy&  Worst Model&Worst Model\\\midrule
         Lorenz&  BD-LSTM&  SMOTER-regular&  Conv-LSTM&no-resampling\\
         Bike&  BD-LSTM&  no-resampling&  BD-LSTM&SMOTER-bin\\
         Sunspot&  Conv-LSTM&  SMOTER-bin&  BD-LSTM&no-resampling\\
 Cyclone-SPO& Conv-LSTM& SMOTER-bin& BD-LSTM&no-resampling\\
 Cyclone-SIO& Conv-LSTM& no-resampling& BD-LSTM&SMOTER-regular\\
 \hline
    \end{tabular}
\end{center}

\caption{Best and worst performing deep learning model combinations with data resampling strategies for extreme value prediction (SER = 1\%).}

\label{tab:summaryprediction}
\end{table*}

 The performance differences among resampling strategies can be partially explained by the distributional characteristics of the target variables.Figure \ref{fig:cyclone} presents the cyclone datasets as an example, and we provide the distribution of bike dataset in Appendix. In both datasets,   no-resampling strategy outperforms synthetic methods and exhibits strongly right-skewed distributions with a sharp decline in sample frequency beyond the extreme thresholds (0.664 and 0.406, respectively), resulting in minimal density in the tail region. This distributional sparsity renders synthetic oversampling prone to generating unrealistic patterns, thereby degrading performance, as observed in Table \ref{tab:summaryprediction}. 

\begin{figure*}[H]
     \centering
     \begin{subfigure}[b]{0.48\textwidth}
         \centering
         \includegraphics[width=\textwidth]{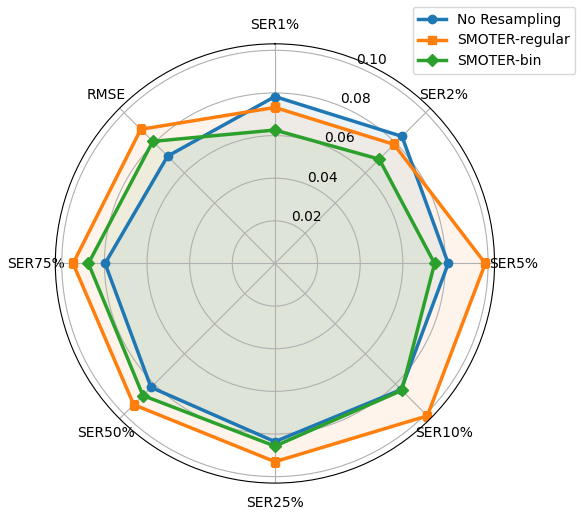}
         \caption{Conv-LSTM}
         \label{fig:y equals x}
     \end{subfigure}
     \begin{subfigure}[b]{0.48\textwidth}
         \centering
         \includegraphics[width=\textwidth]{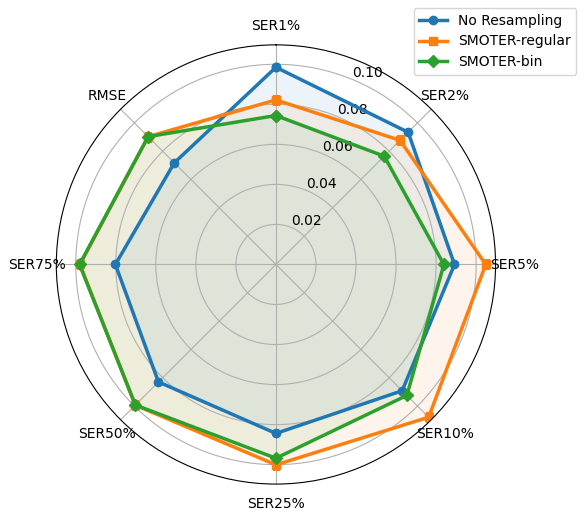}
         \caption{BD-LSTM}
         \label{fig:three sin x}
     \end{subfigure}
        \caption{\textbf{Performance comparison of three resampling strategies on the \textit{Cyclone-SPO} dataset using Conv-LSTM (a) and BD-LSTM (b). } Blue lines represent \textit{No Resampling}, orange lines represent \textit{SMOTER-regular}, and green lines represent \textit{SMOTER-bin}.}
        \label{fig:cyclone radar}
\end{figure*}

\begin{figure*}[H]
     \centering
     \begin{subfigure}[b]{0.48\textwidth}
         \centering
         \includegraphics[width=\textwidth]{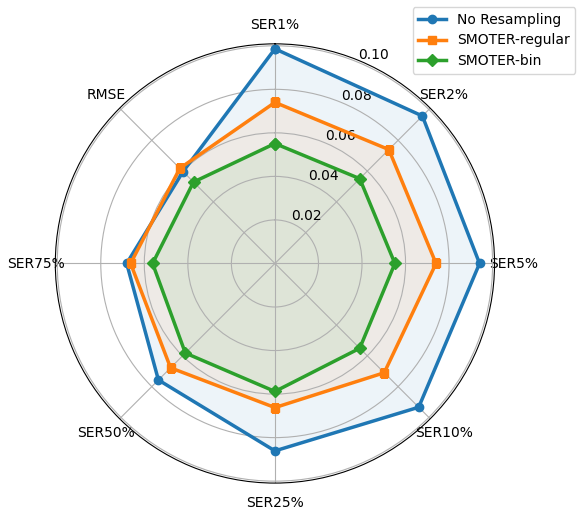}
         \caption{Conv-LSTM}
         \label{fig:y equals x}
     \end{subfigure}
     \begin{subfigure}[b]{0.48\textwidth}
         \centering
         \includegraphics[width=\textwidth]{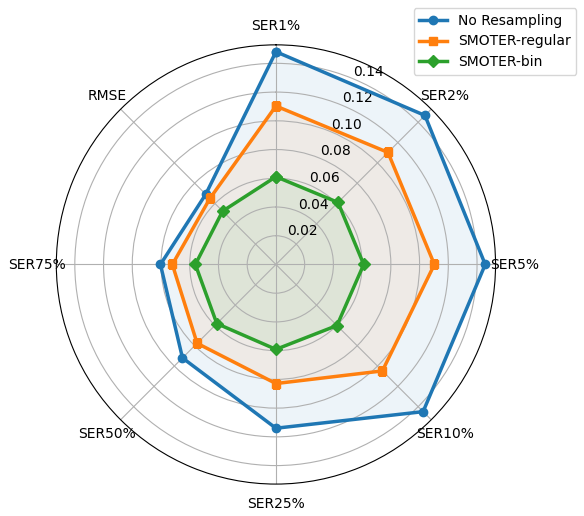}
         \caption{BD-LSTM}
         \label{fig:three sin x}
     \end{subfigure}
        \caption{\textbf{Performance comparison of three resampling strategies on the \textit{Sunspot} dataset using Conv-LSTM (a) and BD-LSTM (b). } Blue lines represent \textit{No Resampling},  orange lines represent \textit{SMOTER-regular}, and  green  lines represent \textit{SMOTER-bin}.}
        \label{fig:sunspot radar}
\end{figure*}

 In such cases, the number of extreme samples is already limited and relatively distinct from the main data mass. Applying methods such as  SMOTER-bin in these contexts risks distorting the original signal and amplifying noise, especially when the extreme region is sparse and volatile. This aligns with the poor performance of SMOTER-bin on the Bike dataset, as shown in Table \ref{tab:summaryprediction}. Figure~\ref{fig:cyclone radar}  and Figure~\ref{fig:sunspot radar} present radar plots comparing resampling strategies across SER thresholds and RMSE values for the Cyclone-SPO and Sunspot datasets. These visualisations help clarify how model–strategy combinations perform under varying levels of evaluation strictness.



In the Cyclone-SPO dataset (Figure~\ref{fig:cyclone radar}), SMOTER-bin achieves the best performance under strict evaluation conditions, particularly from SER1\% to SER10\%, where capturing rare events is most critical. Its error is consistently lower than both SMOTER-regular and no-resampling in this range. However, as the threshold becomes more relaxed (SER25\% and above), the performance of SMOTER-bin degrades, eventually becoming comparable or worse than other methods. This suggests that while SMOTER-bin is highly effective in emphasising rare events, it may over-amplify certain regions when the evaluation shifts toward more frequent values. Interestingly, Conv-LSTM demonstrates better overall stability and lower variance than BD-LSTM in this setting, indicating that the unidirectional structure of Conv-LSTM may be better suited for capturing the moderate regularity present in the Cyclone dataset. SMOTER-regular, by contrast, shows erratic behaviour, with relatively strong performance at SER1\% but significant fluctuations across other thresholds—highlighting its sensitivity to the placement of synthetic samples in irregular temporal contexts.

 By contrast, the Sunspot dataset (Figure~\ref{fig:sunspot radar}) shows a more consistent and favourable response to SMOTER-bin across all thresholds. Both Conv-LSTM and BD-LSTM combined with SMOTER-bin exhibit superior performance, particularly under stricter evaluations. This aligns with the highly periodic nature of the Sunspot series, where SMOTER-bin’s segment-wise oversampling can reinforce important rare patterns without disrupting the underlying signal. The performance gap between SMOTER-bin and the other two strategies is especially pronounced at SER1\%–SER10\%, suggesting that it effectively enhances rare-event representation in datasets with strong cyclical structure. In this case, both models benefit from the clear temporal rhythm of the data, though Conv-LSTM still shows slightly lower variance, likely due to its simpler architecture being better matched to the smooth signal dynamics.
These contrasting patterns highlight how model and strategy effectiveness are shaped by the underlying data properties. For datasets such as Cyclone-SPO, where temporal regularity exists but is less pronounced and more chaotic, simpler models such as Conv-LSTM would generalise better, especially when combined with localised sampling such as  SMOTER-bin. BD-LSTM’s more complex structure may be more prone to overfitting in such scenarios. In contrast, highly regular datasets like Sunspot allow both models to perform well, but benefit most from structure-aware augmentation such as SMOTER-bin. SMOTER-regular, though occasionally competitive, suffers from instability due to its less targeted resampling process.

Overall, the radar plots emphasise that there is no one-size-fits-all solution, SMOTER-bin demonstrates strong potential, particularly under strict evaluation and in datasets with defined temporal patterns. However, its effectiveness depends on both model compatibility and data characteristics. These findings reinforce the need for carefully tailored model–strategy combinations that consider both distributional sparsity and temporal structure when forecasting rare events.

\section{Discussion}

  Our experiments using baseline strategies revealed that SMOTER-bin and SMOTER-regular approaches generally outperformed the GAN-based approaches. SMOTER-bin showed strong short-term gains, particularly at looser thresholds where extreme events are relatively frequent, while SMOTER-regular emerged as the more robust choice under stricter thresholds, maintaining balanced SER and RMSE values. In contrast, both 1D-GAN and 1D-Conv-GAN displayed unstable behaviour, with fluctuating performance that limited their reliability across datasets. The no-resampling strategy, although occasionally competitive in volatile series such as Bike, was consistently surpassed by relevance-guided augmentation, underscoring the effectiveness of SMOTE-based methods in preserving signal fidelity under rare event regimes.

The multi-step forecasting experiments showed that not only does error grow with time—as expected—but that the "best" strategy shifts across steps. This was particularly evident in BD-LSTM (Figure\ref{fig:5 steps}), whose bidirectional nature makes it more sensitive to data augmentation quality. SMOTER-bin remained stable in earlier steps, but could degrade in the long term, depending on the dataset, implying that time is not just an axis in prediction, but a force that amplifies design decisions — highlighting the importance of alignment of strategy, model, and data over time.

Due to the diversity of outcomes across datasets, we evaluated selected deep learning models across five datasets, using a set of  SER thresholds to capture extreme values. The results in Table \ref{tab:DL table} comparing  Conv-LSTM and BD-LSTM models demonstrate that no configuration wins universally. We found that periodic datasets such as Sunspot favoured simpler, unidirectional models with clean augmentation Conv-LSTM.  Moroever, datasets such as Cyclone-SPO, Cyclone-SIO and Lorenz demanded more flexible architectures. The radar plots (Figure \ref{fig:sunspot radar} and Figure \ref{fig:cyclone radar}) visualised these patterns sharply: SMOTER-bin was robust under strict thresholds, while SMOTER-regular wavered, and no-resampling struggled to detect rare events. These insights imply that success in extreme forecasting is less about choosing “the best model” and more about understanding how data, structure, and augmentation interact.

 This study highlights how data augmentation, model architecture, and dataset characteristics interact in shaping deep learning performance for extreme value forecasting, and it also opens new opportunities for exploration. A particularly promising direction is the use of ensemble-based frameworks, where different architectures—such as Conv-LSTM, BD-LSTM, and emerging transformer variants—are combined to harness their complementary strengths. Such ensembles could adaptively adjust their weighting according to relevance scores, forecast horizon sensitivity, or model uncertainty, offering greater resilience in rare-event prediction. Integrating these ideas with adaptive resampling strategies would allow the sampling intensity to evolve alongside the model’s understanding of the data, potentially counteracting the performance drop often seen in long-horizon forecasts. 

 A key limitation of this study lies in the definition of extremes. The identification of rare events relies on a PCHIP-based relevance function combined with discrete thresholds ($\tau = 0.7, 0.8, 0.9$). Although our framework provides a consistent basis for evaluation, the corresponding value thresholds differ markedly across datasets under the same $\tau$, indicating that extreme definitions remain one of the most sensitive and uncertain aspects of the modelling process. Future work could explore adaptive thresholding or more sophisticated relevance functions to capture extremes under dynamic or non-stationary conditions better.  

 Another limitation concerns the scope of generative augmentation. This study examined only 1D-GAN and 1D-Conv-GAN, whose unstable performance may reflect architectural and training constraints rather than an inherent weakness of generative approaches. The potential for extreme value data augmentation using generative models \cite{harshvardhan2020comprehensive} remains largely unexplored, and therefore, novel GANs, diffusion models, and transformer-based generative frameworks can be explored in future work. We note that such models have been mostly used for generating image and video data, and it is essential to evaluate their applicability in generating time series data.

 Future research could embed domain-specific constraints into ensemble systems - for example, incorporating physical laws in environmental forecasting \citep{raissi2019physics} or integrating risk measures and regulatory rules in financial contexts \citep{rockafellar2000optimization}, thereby enhancing both generalisation and interpretability. Such approaches may help build more adaptive and resilient frameworks for extreme value forecasting, capable of evolving with increasing data complexity and the demands of real-world deployment. Another critical frontier is uncertainty quantification. Bayesian deep learning techniques offer promising avenues for projecting predictive uncertainty, for instance, through variational inference \citep{jin2022variational} or MCMC-based sampling schemes \citep{nguyen2023sequential,chandramcmc2024}, which can provide more reliable modelling of tail risks. Beyond improving robustness, these methods also open opportunities for data augmentation strategies—for example, embedding Bayesian structures into SMOTER variants so that sampling intensity adapts dynamically to posterior uncertainty or tail-risk estimates.

\section{Conclusion} 
 This study addressed the persistent challenge of forecasting rare events in time series data by systematically evaluating resampling strategies and introducing a novel deep learning framework. Through extensive experimentation across diverse datasets, model architectures, and evaluation thresholds, we find that model performance is not solely determined by architectural complexity or resampling intensity, but by the alignment of all components—model, data, and augmentation method.

Among the resampling approaches, SMOTER-bin consistently demonstrated superior adaptability, particularly when paired with median quantile forecasting. Its localised sampling mechanism enables better representation of extreme regions while preserving structural integrity, leading to improved performance across both short- and long-horizon forecasts. Conv-LSTM and BD-LSTM exhibit complementary strengths: the former excels in periodic, stable datasets, while the latter performs better in chaotic or non-stationary sequences.

Our results highlight the need for context-sensitive design in extreme value forecasting. Rather than searching for a single optimal strategy, future work should focus on developing adaptive systems that dynamically adjust model and resampling choices based on the statistical and temporal characteristics of the data. This research offers both a conceptual foundation and practical tools for such developments, moving us closer to reliable and interpretable forecasting under distributional uncertainty.

\section*{Code and Data}

Code and data available in our GitHub repo \footnote{\url{https://github.com/sydney-machine-learning/forcastingextremes-dataaugmentation}}

 \newpage
\section*{Appendix}

\begin{figure*}[htbp!]
     \centering
     \begin{subfigure}[b]{1\textwidth}
         \centering
         \includegraphics[width=\textwidth]{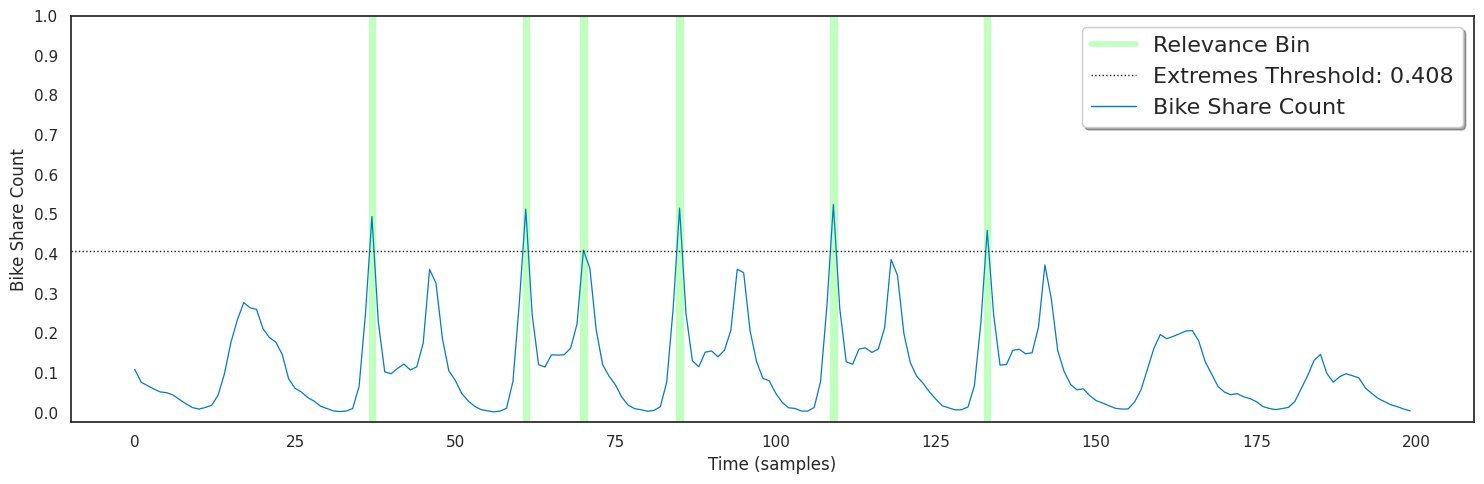}
         \caption{Cyclone time series with bins corresponding to a 0.7 relevance threshold} 
     \end{subfigure}
     \begin{subfigure}[b]{1\textwidth}
         \centering
         \includegraphics[width=\textwidth]{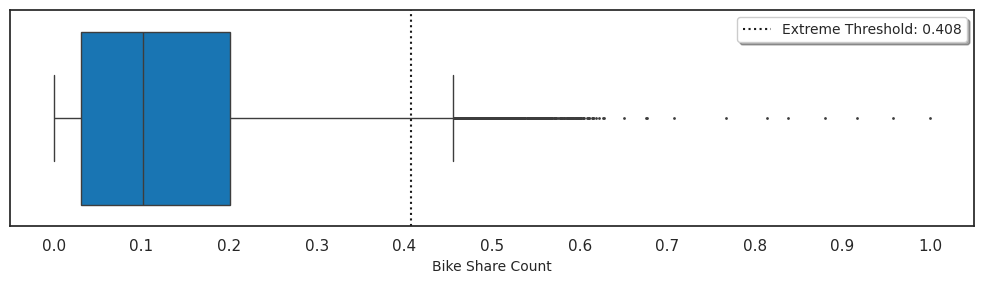}
         \caption{Cyclone boxplot} 
     \end{subfigure}
     \begin{subfigure}[b]{1\textwidth}
         \centering
         \includegraphics[width=\textwidth]{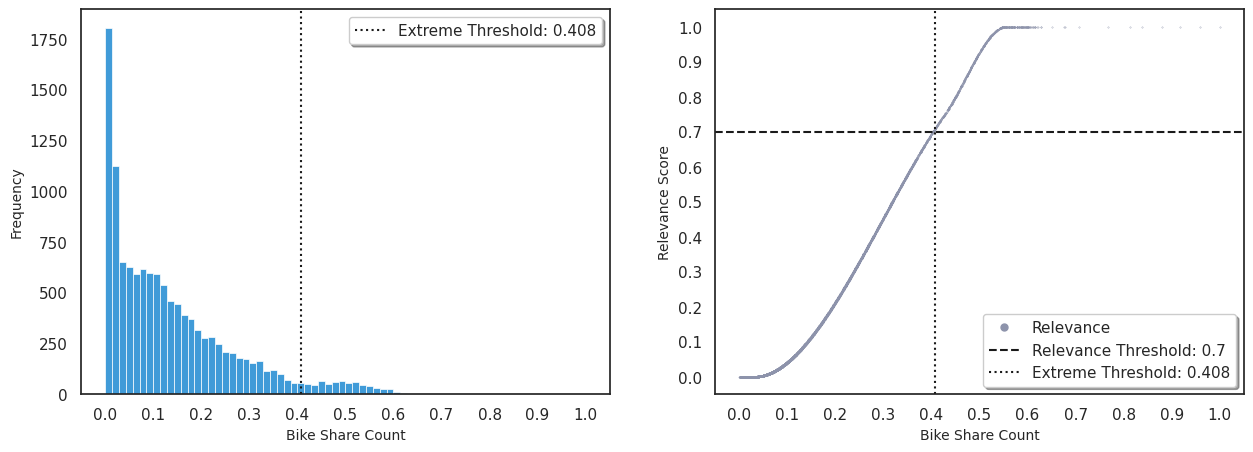}
         \caption{Distribution of Bike time series data as well as the distribution of the extremes at a 0.7 relevance threshold. Also shows the relevance function constructed using PCHIP and the conversion between relevance threshold and extreme threshold.}
         \label{fig:bikerelevance}
     \end{subfigure}
        \caption{Bike dataset visualisation}
        \label{fig:bike}
\end{figure*}

\bibliographystyle{ieeetr}
\bibliography{cas-refs}


\end{document}